\documentclass{article}[10pt]
\usepackage[utf8]{inputenc}
\usepackage{url}
\usepackage{booktabs}
\usepackage{graphicx}
\usepackage{tabularx}
\usepackage{ltablex}
\usepackage{tabu}

\usepackage{tabulary}

\usepackage[newcommands]{ragged2e}
\usepackage{blindtext}
\usepackage{geometry}
\usepackage{titlesec}

\titleformat{\paragraph}
{\normalfont\normalsize\bfseries}{\theparagraph}{1em}{}
\titlespacing*{\paragraph}
{0pt}{3.25ex plus 1ex minus .2ex}{1.5ex plus .2ex}

\begin{document}
 
\title{Recent Advances in Automated Question Answering In Biomedical Domain}
\author{Krishanu Das Baksi, \\School of Information Technology, IIT Delhi; TCS Research\\ krishanudb@cse.iitd.ac.in, krishanu.dasbaksi@tcs.com}
\date{October 2021}
\maketitle

\begin{abstract}
  The objective of automated Question Answering (QA) systems is to provide answers to user queries in a time efficient manner. The answers are usually found in either databases (or knowledge bases) or a collection of documents commonly referred to as the corpus. In the past few decades there has been a proliferation of acquisition of knowledge and consequently there has been an exponential growth in new scientific articles in the field of biomedicine. Therefore, it has become difficult to keep track of all the information in the domain, even for domain experts. With the improvements in commercial search engines, users can type in their queries and get a small set of documents most relevant for answering their query, as well as relevant snippets from the documents in some cases. However, it may be still tedious and time consuming to manually look for the required information or answers. This has necessitated the development of efficient QA systems which aim to find exact and precise answers to user provided natural language questions in the domain of biomedicine. In this paper, we introduce the basic methodologies used for developing general domain QA systems, followed by a thorough investigation of different aspects of biomedical QA systems, including benchmark datasets and several proposed approaches, both using structured databases and collection of texts. We also explore the limitations of current systems and explore potential avenues for further advancement.

\end{abstract}

\section{Introduction}
Gathering information or knowledge is one of the most primary and important step of any logical decision making process. Traditionally, large volumes of information or knowledge was collected and stored in books. Because of high amount of manual effort involved, searching for specific information needs in books and other traditional forms of information storage is difficult and time consuming. However, with the advancement of computer hardware and software, especially technologies for storing and retrieving large volumes of information, computing devices are rapidly becoming the platform of choice for gathering information. 

One of the primary techniques for gathering information is asking questions, a task in which computational systems are increasingly getting better. Question answering by computational systems (QA) is related to but quite different from broader field of information retrieval (IR). Whereas classical IR deals with finding and delivering documents or snippets of documents relevant to a particular query, QA usually aims to find exact answers to a given natural language question \cite{laurent2006qa}. 

Like all fields of science, in the fields of healthcare, medicine and biology, practitioners and scientists have information needs which can be formulated as natural language questions. For e.g. a scientist or a clinician might be interested in knowing "Which enzyme does the drug Tosilizumab target?". The most reliable, precise and up to date information needed to answer these questions are usually contained in peer reviewed scientific research articles, similar to other domains of science. Due to the effort of scientists and clinicians all across the world, our understanding of the world of biology and medicine is mushrooming and large number of scientific papers are being published. Consequently, Pubmed, the biomedical scientific literature database is growing rapidly \cite{VARDAKAS2015592}. This proliferation of articles and knowledge makes it difficult for domain experts to remain up-to-date with the current state of knowledge. In addition, decision making is increasingly becoming dependent on scientific evidence. For e.g., there has been a paradigm shift in the domain of healthcare commonly known as "Evidence-based medicine" \cite{sackett1996evidence}, where clinicians are encouraged to use best evidence from literature for making clinical decisions. The evidence or information need prompts the clinicians to ask questions. Since doctors are typically busy, they need precise and exact answers to their questions very quickly \cite{ely1999analysis}. However, formal literature search is rarely performed \cite{ely1999analysis}, even though biomedical literature can answer a large chunk of the questions (~50\%)\cite{chambliss1996answering}. A likely reason being that a long time (~27 mins per question) is needed to find the answers using search techniques \cite{chambliss1996answering}. As explained before, classic IR systems only return a list of documents which might be relevant for an information need, but the documents need to be read manually to extract the precise relevant information, which is a tedious and time consuming process . The time spent on searching for necessary information can be significantly reduced by developing efficient and accurate biomedical QA systems. Apart from helping in clinical decision making, biomedical question answering systems can also help in educating the general public, who have become increasingly interested in knowing about their own health conditions over the internet \cite{fox2012mobilehealth}.

In this context, automated question answering systems are needed in order to respond to user queries with precise and exact answers. Whereas general domain question answering systems have advanced significantly over the last decade, and now are able to answer natural language questions with reasonable accuracy, because of certain nuances associated with the field of biomedicine (Section \ref{BQAOV}), such QA systems don't work well for biomedical question answering. This has led to growth of interest in biomedical question answering in the natural language processing community and consequently, several datasets and automated QA systems have been proposed.

In this review, several aspects of end-to-end question answering in the biomedical domain will be explored, with a focus on recent advances in question answering, especially methodologies developed in the past decade. We will concentrate on end-to-end QA systems, where only a question is provided, and the objective of the QA system is to return an accurate factoid answer. Unless otherwise stated, the methods described in this review focus on answering factoid questions. Apart from end-to-end factoid QA, Other paradigms of QA like visual QA or conversational QA will not be explored. The paper is organised as follows: In Section \ref{ODQA}, a brief overview of general domain question answering datasets, techniques and some pivotal systems will be discussed. Section \ref{BQAOV} will be contain an overview of the nuances that sets the task of biomedical QA apart from general domain QA. Additionally, different datasets that have been used for training biomedical QA systems will be explored. Section \ref{BQAKB} will comprise of a review of different Biomedical QA systems which use structured knowledge, like knowledge bases. In Section \ref{BQAT}, biomedical QA using text sources will be surveyed. Finally, in Section \ref{DISCUSS}, drawbacks of the current systems will be examined and directions for further progress in the field will be explored. In this review the main focus in on factoid open domain QA systems, although few other types of QA systems will also be discussed.

\section{Question Answering Systems} \label{ODQA}

Question Answering has for long been regarded as one of the most important parameters of judging the "intelligence" of a computational system. During the early phases of NLP and QA research, the primary focus was on the development of rule based systems, i.e. system which process natural language text using hand-crafted linguistic rules or manually identified patterns and produce desired outputs. However, with the development of large datasets and machine learning (ML) techniques, ML based QA systems became more common and currently outperform rule based models. 

In this review the emphasis is on open domain question answering (ODQA). Open domain question answering refers to the task of predicting an answer when only a question without a context is provided. We use the terms "open domain question answering" and "end-to-end question answering" interchangeably. Since in most cases questions cannot be directly processed to produce the answers, end-to-end QA systems usually depend on some kind of a knowledge or information source using which the required answers can be produced. For e.g., in the field of biology this information may be available in unstructured text documents like biological literature \cite{VARDAKAS2015592} or structured databases like Uniprot \cite{uniprot2021}, Gene Ontology \cite{go2021} etc. Based on the source of knowledge used, QA systems can be broadly be categorized into:
\begin{itemize}
    \item Question Answering using Structured Knowledge Bases (KB)
    \item Question Answering using Unstructured Text
    \item Question Answering using Heterogeneous Sources (both text and KB)
    
\end{itemize}
Both of these paradigms will be explained in the following subsections.
It must be noted that some recently described QA systems, especially transformer based models like GPT \cite{gpt3} and T5 \cite{2020t5}, are able to answer questions without using any external information. The parameters of the models (numbering up to hundreds of billions) are believed to contain the knowledge needed to answer any question, although this "parametric knowledge" \cite{lewisRetrievalAugmentedGenerationKnowledgeIntensive} cannot be interpreted by humans.

\subsection{Question Answering using Knowledge Bases}\label{KGQA}
In this paradigm, the systems try to answer natural language questions using structured knowledge bases. Large amounts of information can be stored in structured databases that can be queried using specialized formal query language. However, these sophisticated and specialized querying languages are not easy to learn and hence not suitable for all users. Therefore, "natural language front ends to the databases" were designed \cite{hirschman2001natural} to convert questions asked in natural language to a formal query using different techniques. The formal query could be used to extract answers from the database. BASEBALL \cite{baseball61green}, LUNAR \cite{woods1972lunar}, RENDEZVOUS \cite{codd1974seven}, LADDER \cite{hendrix1978ladder}, Chat-80 \cite{warren1982chat80}, MASQUE \cite{auxerre1986masque1, auxerre1986masque2} are some of the early systems of this paradigm. These systems relied on hand crafted linguistic rules to convert natural language questions to formal queries. Since linguistic rules are difficult to generalize to new domains, these early systems suffered from several drawbacks like narrow domain specificity, lack of portability, etc. 

Knowledge graphs (KG) or semantic networks (Section \ref{KGintro}) are a special type of database where the data can be represented in the form of a graph. In the last decade and a half, several knowledge graphs in different domains, including biomedicine, have been constructed (also referred to as knowledge bases). Because of their inherent structure (which can be easily processed by computers) and the breadth of information they contain, KGs  have been used in several question answering systems. Currently KGs are the most common form of structured knowledge used for QA. QA systems which use KGs as the source of knowledge or information are generally referred to as KGQA methods.

\begin{table}[htbp]
\centering
\caption{Properties of common datasets used for benchmarking KGQA systems. For all the benchmarks, the questions are single sentences and both the questions and answers are in English language}
\label{tab:benchmarks}
\resizebox{\textwidth}{!}{%
\begin{tabular}{cccccll}
\hline
\textbf{Name of Dataset} & \textbf{Question Complexity} & \textbf{Answer Source}                                               & \textbf{Formal Query}                                                                              & \textbf{\begin{tabular}[c]{@{}c@{}}Number of \\ QA pairs\end{tabular}} & \textbf{Type of Questions} & \textbf{Evaluation Metric}                                               \\ \hline
WebQuestions  \cite{berant2013semantic}           & Simple and Complex           & KG (Freebase \cite{bollacker2007freebase})                                                        & No                                                                                                 & 5810                                                                   & Factoid and List           & Accuracy, F1 score                                                       \\ \hline
WebQuestionsSP \cite{yih2016value}           & Simple and Complex           & KG (Freebase)                                                        & Yes. In SPARQL                                                                                     & 4737                                                                   & Factoid and List           & F1 Score                                                                 \\ \hline
ComplexQuestions \cite{bao2016constraint}        & Simple and Complex           & KG (Freebase)                                                        & No                                                                                                 & 2100                                                                   & Factoid and List           & F1 Score                                                                 \\ \hline
SimpleQuestions \cite{bordes2015large}         & Simple                       & KG (Freebase)                                                        & \begin{tabular}[c]{@{}c@{}}KG triple capable of \\ answering question \\ is  provided\end{tabular} & 100000+                                                                & Factoid                    & Accuracy                                                                 \\ \hline
LcQUAD \cite{trivedi2017lc}                  & Simple and Complex           & KG (DBPedia \cite{dbpedia})                                                         & Yes. In SPARQL                                                                                     & 5000                                                                   & Yes/No, Factoid, List      & \begin{tabular}[c]{@{}l@{}}Exact Match Accuracy,\\ F1 score\end{tabular} \\ \hline
LcQUAD 2 \cite{dubey2019lc}                & Simple and Complex           & \begin{tabular}[c]{@{}c@{}}KG (DBPedia \\ and Wikidata \cite{vrandevcic2014wikidata})\end{tabular} & Yes. In SPARQL                                                                                     & 30000                                                                  & Yes/No, Factoid, List      & \begin{tabular}[c]{@{}l@{}}Exact Match Accuracy,\\ F1 score\end{tabular} \\ \hline

\end{tabular}%
}
\end{table}

\subsubsection{Knowledge Graphs} \label{KGintro}
Knowledge graphs (KGs) or semantic networks have recently gained the interest of industry and academia as a form of structured knowledge. A knowledge graph is a structured graphical representation of facts and relations among entities along with semantic descriptions of the entities. Entities (nodes) can represent real world objects or abstract concepts, whereas relations (edges) represent the relationship among entities, and semantic descriptions denotes the properties of entities and relationships. In a KG, knowledge or facts are represented as knowledge triples, where each triple is of the form of (subject, predicate, object), adhering to the Resource Description Framework (RDF). For e.g. (Glucagon, affects, Hepatitis A) is an RDF triple where "Glucagon" (subject) and "Hepatitis A" (object) are the entities, and "affects" is the relation (directed edge) among the entities. An example KG is depicted in Figure 1. Knowledge graphs can be constructed using both manual curation, for example Yago \cite{fabian2007yago}, DBPedia \cite{dbpedia} etc., or using automated information extraction (IE) techniques like NELL \cite{carlson2010toward} and OpenIE \cite{etzioni2011open}. Construction of KGs by manual curation is tedious and time consuming. Additionally, since it is difficult to explicitly curate all facts, for e.g. common sense facts, therefore the manually curated KGs are notoriously incomplete \cite{jianhui2018research, kgcompletionreview2020} i.e. have low recall. Several techniques for knowledge graph completion i.e. predicting new relationships among entities in the KG using the known relationships have been described \cite{kgcompletionreview2020}. IE methods can potentially make KG creation easier, but in practice, the automatically created KGs are typically very noisy and not frequently used for QA. For a comprehensive review of knowledge graphs and different associated tasks, readers can refer to the following reviews \cite{ji2021survey}.

Some of the well known manually curated general domain knowledge graphs are YAGO \cite{fabian2007yago}, DBPedia \cite{dbpedia}, Freebase \cite{bollacker2007freebase} and Wikidata \cite{vrandevcic2014wikidata}. These KGs are usually very large and contain a plethora of information. For e.g., Wikidata, currently the largest and most popular general domain KG, has 12 billion facts, 84 million entities, 7 thousand predicates and 63 thousand entity types. Specialized query language are also designed to query knowledge graphs, the most popular among them being SPARQL \cite{SPARQLQueryLanguage}. Since KGs contain a plethora of information in a structured or computer readable form, they have been used for constructing question answering systems. A majority of KGQA systems described in literature use manually curated KGs like Freebase \cite{bollacker2007freebase}, although methods which answer questions using automatically curated KGs have also been developed \cite{fader-etal-2013-paraphrase, 10.1145/2623330.2623677}.

\begin{figure}[]
    \label{fig:examplekg}
    \centering
    \includegraphics[width=\linewidth]{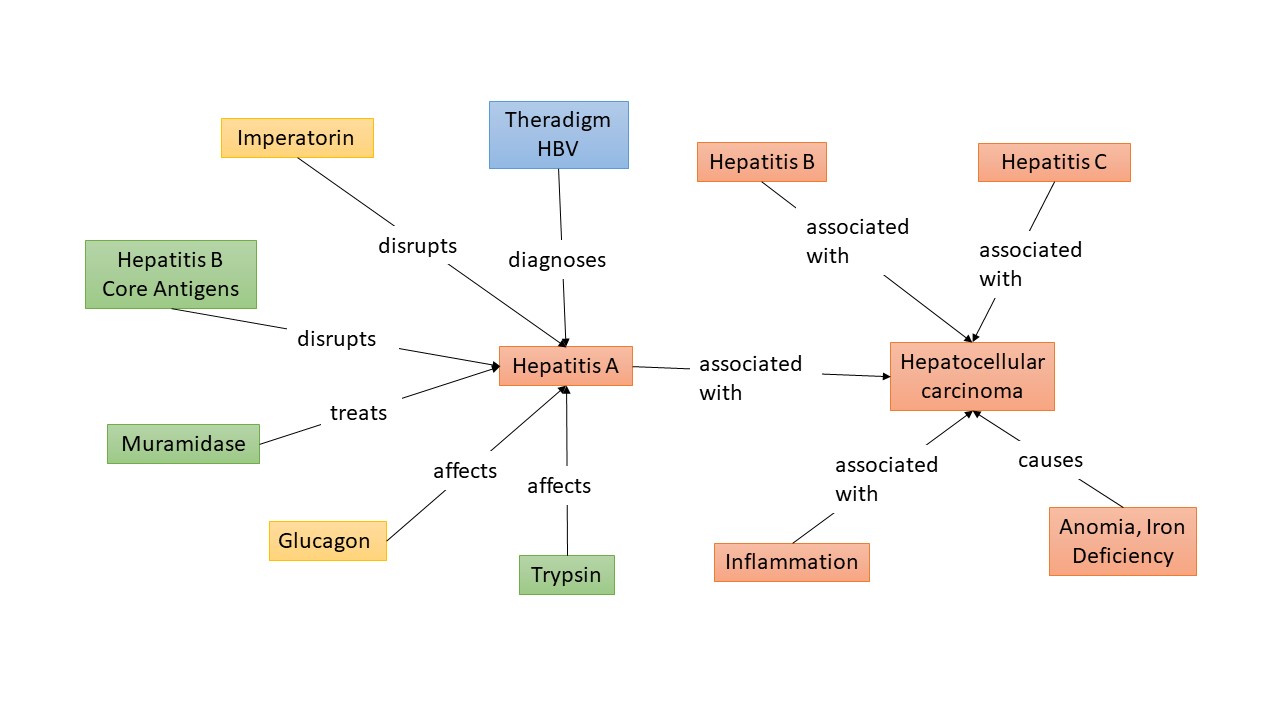}
    \caption{A small part of the biomedical Knowledge Graph in KGHC (Knowledge Graph for Hepatocellular Carcinoma \cite{li2020kghc}), a narrow biomedical KG pertaining to hepatocellular diseases. Here, the different edges represent the relationships among different entities. In order to reduce clutter in the figure, the different node colors are used to represent entity types, e.g., Orange nodes depict type "Disease", Yellow nodes represent type "Hormone", etc.}
\end{figure}

\subsubsection{Benchmarks}
With the development of KGQA methods, benchmark datasets became necessary for the evaluation and comparison of the performance of the different methods. A summary of the most commonly used benchmark datasets used in QA using knowledge bases as well as text corpora is tabulated in Table \ref{tab:benchmarks}

One way to classify questions over knowledge graphs is using the number of relationship triples used for answering the questions. Following this methodology, there can be two types of questions:
\begin{enumerate}
    \item \textbf{Simple questions} (not to be confused with the benchmark dataset SimpleQuestions \cite{bordes2015large}) are those which can be answered using a single fact or triple in the KG. For e.g. in the examle KG depicted in Figure 1, "What diagnoses Hepatitis A?" is a simple question, as it involves only one relationship triple ("Theradigm HBV", "diagnoses", "Hepatitis A").
    
    \item One the other hand \textbf{complex questions} need more than one relationship triple to be answered correctly, and typically include operations such as chaining, aggregation, conjunction or disjunction of multiple relationship triples. For e.g. in the example KG depicted in Figure 1, "What is used to treat the disease diagnosed by Theradigm HBV?" is a complex question, as it requires chaining of two relationship triples ("Theradigm HBV", "diagnoses", "Hepatitis A") and ("Muramidase", "treats", "Hepatitis A") in order to answer the question correctly.
\end{enumerate} 

Some of the popular benchmark datasets for evaluation of KG question answering systems are listed below.  These benchmarks are either manually generated or computationally generated. A general property of datasets generated manually is that they are comparatively smaller in size, but are more realistic since it is difficult for machine learning models to learn. On the other hand automated or semi-automated methods of dataset construction usually leads to much larger datasets, but since the data (questions and/or answers and/or queries) are extracted using patterns and neural models are very good at learning those patterns. Therefore, neural models perform very well on these datasets even without learning the ability to generalize and answer all types of questions.

\begin{enumerate}
    \item SimpleQuestions \cite{bordes2015large} (based on Freebase) is one of the largest manually curated knowledge based question answering datasets containing around one hundred thousand data points, where each data point contains the question, answer and the relationship triple that can be used to derive the answer. It was created over Freebase and consists of simple questions. 
    
    \item WebQuestions \cite{berant2013semantic} (based on Freebase) consists of questions fetched from Google search API and answers to those questions provided by crowdsourced workers. However, the original dataset does not contain the formal queries to find the answers. WebQuestionsSP \cite{yih2016value} contains the annotated SPARQL queries for the questions in WebQuestions, whereas some unclear and ambiguous questions are removed. A majority (84\% \cite{fu2020survey}) of the questions in WebQuestions are simple, i.e. can be answered using a one hop query or one relationship triple. However, often, more complex or multihop questions are asked. In order to evaluate QA systems' performance in answering multihop questions, ComplexQuestions \cite{bao2016constraint} was developed. In this dataset, questions with type, temporal and aggregation constraints are added to the WebQuestions dataset.
    
    \item Large Scale Question Answering Dataset or LcQUAD (based on DBPedia) \cite{trivedi2017lc} consists of a majority (82\%) complex questions. Predefined SPARQL query templates were filled with specific entities and relations to generate a complete query, which was later turned into natural language questions using predefined question templates and crowdsourcing. Following a similar methodology, LcQUAD 2.0 (based on both Wikidata and DBPedia) \cite{dubey2019lc} was created, which contains more diverse types of complex questions.
    
    \item Several other datasets have been described in literature viz, ComQA \cite{abujabal-etal-2019-comqa}, GraphQuestions \cite{su-etal-2016-generating}, QALD, TempQuestions \cite{tempqa} and WikiMovies \cite{miller-etal-2016-key}.
\end{enumerate}

\subsubsection{Methods for QA}
Based on the methodology, systems for answering questions using knowledge graphs can be classified into the following categories:
\begin{enumerate}
    \item Template based methods
    \item Query construction based methods
    \item KG embedding based methods
\end{enumerate} 

We briefly describe the core ideas of some of the seminal work in each of these three categories of methods.

\paragraph{Template based Methods for Question Answering over Knowledge Graphs}

\begin{table}[]
\centering
\caption{Representative examples of different types of KGQA systems.}
\label{tab:kgqa}
\resizebox{\textwidth}{!}{
\begin{tabular}{llllll}
\hline
\textbf{Method} & \textbf{Knowledge Graph} & \textbf{Question Complexity} & \textbf{Method Classification}                                                       \\ \hline
QUINT \cite{quint17}                      & Freebase                                                                                          & Simple and Complex           & Template Based                                                                       \\ \hline
PARALEX \cite{fader-etal-2013-paraphrase} & \begin{tabular}[c]{@{}l@{}}Custom KG created \\ using open information \\ extraction\end{tabular} & Simple                       & Template Based                                                                       \\ \hline
SEMPRE \cite{berant2013semantic}          & Freebase                                                                                          & Simple and Complex           & \begin{tabular}[c]{@{}l@{}}Query construction \\ using semantic parsing\end{tabular} \\ \hline
STAGG \cite{yih-etal-2015-semantic}       & Freebase                                                                                          & Simple and Complex           & \begin{tabular}[c]{@{}l@{}}Query construction \\ using neural networks\end{tabular}  \\ \hline
QAmp \cite{10.1145/3357384.3358026}       & DBPedia                                                                                           & Simple and Complex           & \begin{tabular}[c]{@{}l@{}}Query construction \\ using neural networks\end{tabular}  \\ \hline
KEQA \cite{10.1145/3289600.3290956}       & Freebase                                                                                          & Simple                       & \begin{tabular}[c]{@{}l@{}}KG Embedding based \\ method\end{tabular}                 \\ \hline
EmbedKGQA \cite{embedkgqa}                & \begin{tabular}[c]{@{}l@{}}Freebase; \\ MetaQA-KG \cite{metaqa}\end{tabular}     & Simple and Complex           & \begin{tabular}[c]{@{}l@{}}KG Embedding based \\ method\end{tabular}                 \\ \hline

\end{tabular}
}
\end{table}

\begin{figure}[h]
  \centering
  \label{fig:templateeg}
  \includegraphics[width=\linewidth]{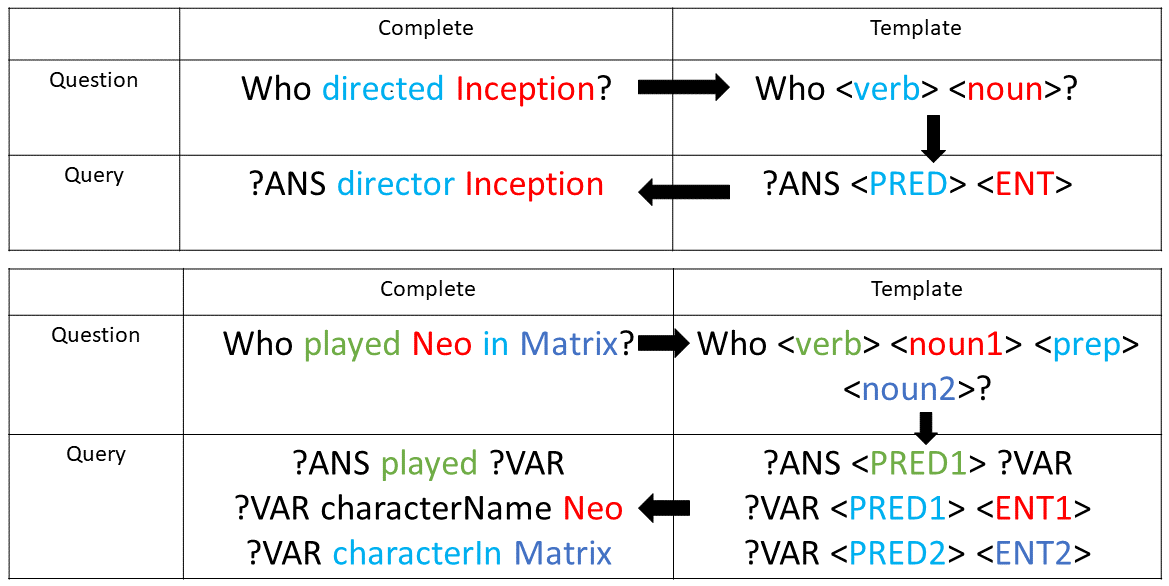}
  \caption{Examples of template based QA over KGs. The original questions are converted to a question template via semantic and/or syntactic parsing. The question template is mapped to the corresponding query template. The question surface forms are mapped to KB entries to fill the variables/ placeholders in the query template to give the final query. The top example is a single hop / relation type question, whereas the bottom example uses multiple relations.}
\end{figure}

Template matching is one of the earliest methods for question answering using knowledge graphs. In this class of techniques, a dataset of question templates and matching query templates are used. A new question is processed using linguistic rules and converted to the question template. The question template is mapped to a query template. Both the question and query templates contain placeholders for different entities and relations. Finally the structured formal query is produced by mapping the surface forms or terms in the question to the knowledge base entries (entities, relations, classes etc.) and using the KG entries to populate the placeholders. The final query is then executed on the knowledge graph and the answer is returned. Examples of the process can be found in Figure 2. The initial focus of this paradigm was on converting simple natural language questions into a triple representation. However, since all questions cannot be represented only using single triples, mechanisms for the conversion of complex questions to formal queries were explored in \cite{10.1145/2187836.2187923}. Since the templates are hand crafted by experts and hence have restricted coverage, the method can be very fragile when unseen question formulations are used \cite{quint17}. One solution to this problem was proposed in the QUINT \cite{quint17, 10.1145/3038912.3052583} framework, which learns the query paths using distant supervision. The question terms and the query entities are aligned using integer linear programming and further processed to create question-query template pairs with placeholders for terms and entities. These templates are used to map new questions to their corresponding templates and entity linking step is used to populate the query template with entities.
The QUINT framework mainly uses syntactic similarity to match new questions to question templates. It cannot be used to answer questions which do not match any known template. The NEQA framework \cite{10.1145/3178876.3186004} expands upon the ideas of QUINT and uses the semantic similarity of new questions (which do not map to any known template) with known questions (in the training set) to derive the most likely query template. The semantic similarity score between a known question and a new question is calculated using: 1. the surface level matches or the word level matches among the questions, 2. the cosine similarity between the word2vec embeddings of different pairs of phrases and words in the questions and 3. the expected answer type of the questions. The template of the highest scoring known question is used to construct the query corresponding to the new question. If the generated query can answer the question correctly (based on feedback from the user), the new question and query pair is added to the known set and the query ranking model is updated. Therefore, the NEQA system is designed to continuously keep improving using positive and negative user feedback.

Whereas the above methods are focused on QA over manually curated and therefore well structured KGs, another line of work focused on extracted KGs, i.e. KGs produced by open information extraction systems viz. NELL \cite{carlson2010toward}, OpenIE \cite{etzioni2008open, etzioni2011open} etc. Compared to manually curated KGs which are typically incomplete but have high precision, extracted KGs are noisy, and entities and relations are not normalized. PARALEX \cite{fader-etal-2013-paraphrase} is a method to answer single relation or single hop questions over extracted KGs. The method derives queries from questions by first matching the question to a template query and then matching the surface forms to the KG entities. It leverages a noisy semi-labelled dataset of questions and their paraphrases or reformulations and a small seed lexicon matching question terms to KG entries to learn the mapping from questions' surface forms to query templates and KG entries. Finally, all the queries derived are ranked and the best ranked query is used to derive the answer. 

Template based methods are inherently easy and interpretable. However as the questions become more complex, the number of possible templates becomes intractable and hence becomes a limiting factor to the performance of this class of methods. 

\paragraph{Methods for Query Construction for Question Answering over Knowledge Graphs}
This class of methods use the semantics of the question to construct a formal query which is executed over the knowledge graph to get the answer. Some of the early work uses semantic parsing to directly construct formal queries from questions, whereas more recent research has focused on learning neural query generation models.

SEMPRE \cite{berant2013semantic} constructs queries that can be executed over the KG using semantic parsing of the input questions. Using a lexicon to map question surface terms to KB entities or relations and a set of composition rules, it recursively constructs more and more complex formal queries from simpler contiguous forms. The system uses a machine learning model to select the best possible derivation (partial formal query) from the possible ones. The model is trained using distant supervision, using only question answer pairs, not using the true formal queries. Although distant supervision eliminates the requirement of manually curated semantic parse of questions, the true semantic parse of the question is not known and only a likely approximation to the true parse is used for training. This makes the learning step less reliable. 

Neural Networks have also been used for query construction. One of the earliest systems for KGQA employing neural networks in its pipeline is STAGG \cite{yih-etal-2015-semantic}. 
In brief, the method identifies the relevant entities in the KG using S-MART \cite{yang-chang-2015-mart} and iteratively predicts the path  connecting the selected entity node to answer node using Siamese Convolutional Neural Networks \cite{bromley1993signature} and finally constraints are added using several rules to construct the query. 

In \cite{10.1145/3357384.3358026}, the authors describe a system named QAmp for answering complex multihop questions. The method breaks down the task of query formation into two steps. In the question interpretation step, the entity, relation and class keywords are first identified from the question surface form using a BiLSTM-CRF \cite{bilstm-crf} model, followed by a reranking step which ranks different terms from the KG matching the identified question surface terms. Several potential query subgraphs are constructed using message passing methodology, and executed to produce several probable answers. The answers are ranked using the aggregated reranking score of different components of the subgraph. Additionally, a prediction of the type of the question (select, count, ask) is made, which helps derive the final answer.


Some methods try to generate the formal query to produce the answer directly using neural networks. In \cite{directq2sg}, a recurrent neural network model directly predicts the head and relation entity for simple questions. Using the head and relation entitites, the answer entity can be predicted. Other methods describing the direct generation of formal queries (in SPARQL) using neural network architectures have also been described \cite{soru2020sparql, wang2020answering}.


\paragraph{Embedding Based Methods for Question Answering over Knowledge Graphs} \label{kgembed}
In this part, we discuss methods which computes embeddings for predicting the answers, thereby not using the formal query to find the answers from the KG. Some of these methods can be trained using only question-answer pairs. Therefore the need for construction of manually curated queries corresponding to the natural language questions, which is a time consuming process, is bypassed \cite{10.5555/3120260.3120272}. 

One of the earliest methods of this type was described in \cite{10.5555/3120260.3120272, bordes-etal-2014-question}. The method first identifies the focus entity of the question and generates answer candidates from the one or two hop neighbours of the focus entity. The method uses two learnable transformation matrices to embed the sparse bag of words representation of question and possible answers/answer paths into a dense multidimensional embedding space, such that the dot product of the embeddings of the questions and candidate answers are high if the answer is the true answer to the question and low otherwise. The transformations matrices used for embedding questions and candidate answers are learned using gradient descent.


KEQA \cite{10.1145/3289600.3290956} uses knowledge graph embeddings as well as deep learning methodologies to predict the answer to simple one hop questions. The embeddings of the entities and relationships in the KG is derived using a TransE \cite{NIPS2013_1cecc7a7} model. Recurrent neural networks (LSTMs \cite{lstm}) are used to predict the embeddings of the main entity and relation in the question, such that they are similar to their corresponding embeddings produced by the TransE model. Going by the principle of translational knowledge embedding models \cite{NIPS2013_1cecc7a7}, the embedding of the answer entity can be predicted using the sum of the predicted head and relationship embeddings. The answer entity can be predicted by finding the entity whose embedding is closest to the predicted answer embedding. This method is tailor made for answering simple questions or one hop questions, and can potentially handle missing links/relations in knowledge graphs. However, it fails on complex or multi hop questions.

In order to answer multihop questions using KG embeddings, methods such as EmbedKGQA \cite{embedkgqa} have been developed. EmbedKGQA, similar to KEQA, finds embeddings (in complex dense multidimensional space) of nodes and relations in the KG using ComplEx \cite{complex}. It finds embeddings (in complex dense multidimensional space) of the questions using RoBERTa \cite{roberta}, a transformer based language model, and a feed forward neural network. The KG embedding model and the question embedding model are trained simultaneously such that a scoring function, viz. the complex scoring function \cite{complex} is positive for correct question, topic entity and answer entity triples, and negative otherwise. During inference, given a question and topic entity, the entity which maximizes the complex scoring function is predicted as the answer. The EmbedKGQA methodology has been shown to be effective on multihop questions of different hop lengths and also on incomplete KGs i.e. KGs with high number of missing links/relations. However, one major drawback is that it requires the question topic entity to be provided with the question, which is highly unlikely to be satisfied in practical settings.

\subsection{Question Answering using Text}

\begin{table}[]
\caption{Properties of common datasets used for benchmarking text based QA systems}
\label{tab:benchmarkstext}
\resizebox{\textwidth}{!}{%
\begin{tabular}{lllllll}
\hline
\textbf{\begin{tabular}[c]{@{}l@{}}Name of \\ Dataset\end{tabular}} & \textbf{Size} & \textbf{Type} & \textbf{Question Type} & \textbf{Answer Type} & \textbf{Remarks} & \textbf{Text Source} \\ \hline
SQuAD 1.0 \cite{rajpurkarSQuAD1000002016} & 100000+ & \begin{tabular}[c]{@{}l@{}}Machine Reading \\ Comprehension\end{tabular} & \begin{tabular}[c]{@{}l@{}}Question and \\ Relevant Passage \\(context)\end{tabular} & Factoid Answer & \begin{tabular}[c]{@{}l@{}}Can be converted to open \\ domain factoid QA by \\ removing the context\end{tabular} & Wikipedia \\ \hline
SQuAD 2.0 \cite{rajpurkarKnowWhatYou2018} & 150000+ & \begin{tabular}[c]{@{}l@{}}Machine Reading \\ Comprehension\end{tabular} & \begin{tabular}[c]{@{}l@{}}Question and \\ Relevant Passage \\ (context)\end{tabular} & Factoid Answer & \begin{tabular}[c]{@{}l@{}}Can be converted to open \\ domain factoid QA by \\ removing the context. \\ Also contains questions \\ which cannot be answered \\ using the context.\end{tabular} & Wikipedia \\ \hline
\begin{tabular}[c]{@{}l@{}}Natural-\\ Questions \cite{kwiatkowski2019natural}\end{tabular} & 300000+ & Open Domain QA & Question Only & \begin{tabular}[c]{@{}l@{}}Factoid Answer, \\ Summary Answer\end{tabular} & \begin{tabular}[c]{@{}l@{}}Can be used to benchmark \\ both long (summary) and \\ short (factoid) \\ type QA systems.\end{tabular} & Wikipedia \\ \hline
\begin{tabular}[c]{@{}l@{}}Curated-\\ TREC \cite{Baudis2015ModelingOT} \end{tabular} & 2180 & Open Domain QA & Question Only & Factoid Answer &  & \begin{tabular}[c]{@{}l@{}}Unspecified \\ but focused \\ on \\ Wikipedia\end{tabular} \\ \hline
\end{tabular}%
}
\end{table}

In the task of end-to-end question answering or open domain question answering using text, the objective is to predict answer to a given question. Usually ODQA systems can be divided into two subsystems: retriever and reader. Retrievers are responsible for document retrieval, i.e. finding documents or text passages from the entire corpus which contain knowledge relevant for answering the given question. On the other hand readers deal with predicting the correct answer given the question and retrieved texts. The task of the reader in ODQA is overlapping with but in essence different from the related task of machine reading comprehension (MRC). In MRC, a context passage and a question that can be answered (usually) using the knowledge in the provided context is provided, and the objective is to predict the answer. In ODQA readers, the contexts are provided by the retriever, therefore may or may not contain the answer.

QA using text is inherently difficult as it requires understanding of unstructured natural language text by models. Consequently, the accuracy of text based QA systems was low before the advent of deep learning. However, recent advances in NLP using deep learning models have significantly improved the accuracy of QA systems using text and several methodologies for this paradigm has been developed in recent years.

The choice of corpus is very crucial for QA using text. Some systems have been designed to answer questions using text from the entire web \cite{brill-etal-2002-analysis}. However, using the entire web as a corpus is problematic due to a number of reasons viz. lack of credibility of web pages, the large size of the corpus and the presence of several languages. Therefore a controlled text corpus is preferred. The Wikipedia corpus is a reasonably trustworthy, manually annotated, up-to-date corpus which contains knowledge about several topics which are of importance to humans. Therefore, the Wikipedia corpus has been often used in designing text based QA systems \cite{chen-etal-2017-reading, karpukhin-etal-2020-dense, izacardDistillingKnowledgeReader2020, izacardLeveragingPassageRetrieval2020, lewisRetrievalAugmentedGenerationKnowledgeIntensive, guuREALMRetrievalAugmentedLanguage2020}. 

The overall methodology of end to end question answering using text can be summarised in Figure 3. The large majority of methodologies developed for the task follow the Retriever-Reader architecture. As mentioned before, in this paradigm, the overall process can be divided into two steps:
\begin{enumerate}
    \item \textbf{Retriever}: The first step is document retrieval, which deals with finding a set of passages or documents from the entire corpus which are relevant for answering the question.
    
    \item \textbf{Reader}: The second step is that of machine reading, which deals with finding the exact answers from the set of relevant retrieved documents.
\end{enumerate}

\subsubsection{Benchmarks}
Datasets used for benchmarking knowledge graph question answering (KGQA) methods can be used for evaluating systems for QA using text, e.g., WebQuestions \cite{berant2013semantic}. Several other benchmark datasets not specific to any knowledge base have also been developed. Some of the datasets comprise of natural language questions and the final answers without information regarding the passage or document from which the answer can be extracted \cite{Baudis2015ModelingOT}. Other are machine reading comprehension datasets, which comprise of questions, answers as well as a context passage from where answers can be extracted \cite{joshi-etal-2017-triviaqa, rajpurkarKnowWhatYou2018, rajpurkarSQuAD1000002016}.

In recent years, the most commonly used dataset for evaluating QA systems is the Natural Questions dataset \cite{kwiatkowski2019natural}. Each question is accompanied by a relevant wikipedia document, a paragraph which can be regarded as the long answer to the question as well as the exact short answer. The questions are derived from Google search queries whereas the answers are manually annotated. It contains a total of approximately three hundred thousand annotated questions, thereby making it one of the largest of its kind.

Some of the most popular question answering benchamark datasets are the SQuAD datasets: SQuAD1.1 \cite{rajpurkarSQuAD1000002016} and SQuAD2.0\cite{rajpurkarKnowWhatYou2018}. The SQuAD datasets are essentially machine comprehension datasets comprised of a question, a context passage from wikipedia and an answer. In SQuAD1.1 \cite{rajpurkarSQuAD1000002016}, all the questions can be answered with the information in the passage, whereas in SQuAD2.0 \cite{rajpurkarKnowWhatYou2018} a third of the questions cannot be answered with the provided passages and hence is more challenging for a ML model to learn. The SQuAD datasets can be used in the open domain QA setting (end to end question answering) by only providing the question as input, to which computational methods must retrieve the relevant paragraph as well as predict the current answer. However, it is important to note that the questions of the SQuAD dataset may lack relevance without the passage that is provided \cite{karpukhin-etal-2020-dense}. During the annotation phase for SQuAD, the annotators were shown the passages and asked to formulate the questions. Therefore, while framing the questions, the annotators often unintentionally used several terms present in the paragraph, thereby making the paragraph and question lexically similar \cite{karpukhin-etal-2020-dense}. This makes it biased towards lexical similarity based retrieval models like BM25 \cite{robertson2009probabilistic}.  

CuratedTREC \cite{Baudis2015ModelingOT} is another benchmark dataset intended for QA over unstructured corpora. It comprises of around two thousand questions sourced from TREC QA tasks.

The WikiMovies \cite{miller-etal-2016-key} dataset comprises of 96 thousand questions and answers in the domain of movies sourced from OMDB and MovieLens databases. It is interesting to note that the questions are designed such that be answered both using knowledge bases as well as a subset of Wikipedia pages, therefore can be used to compare KGQA and text based QA methodologies. However, due to its narrow domain, its applicability in benchmarking general QA systems is limited.

\begin{figure}[h]
  \centering
  \label{fig:text_qa}
  \includegraphics[width=\linewidth]{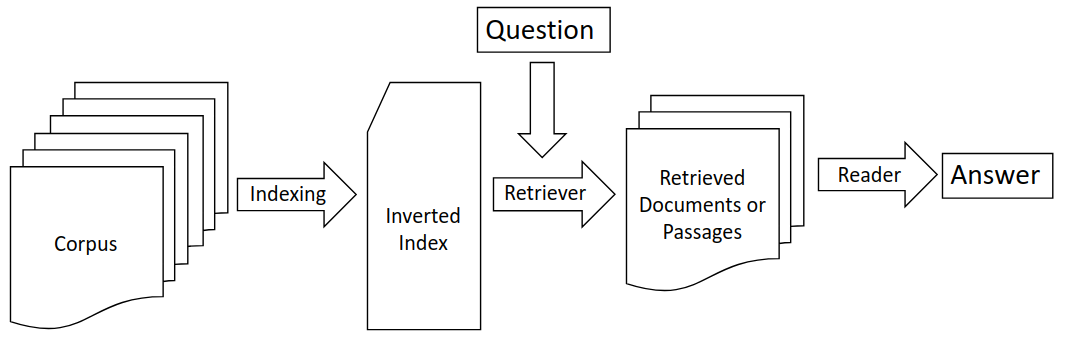}
  \caption{General architecture of text based end to end question answering systems \cite{jurafsky2008speech}}
\end{figure}

\subsubsection{Systems}

\begin{table}[]
\caption{Recent methods for end-to-end factoid QA using Text. All the listed systems use Wikipedia as the text source.}
\begin{tabularx}{\textwidth}{XXX}
\hline
\textbf{Method} & \textbf{Retriver} & \textbf{Reader} \\\hline
DrQA \cite{chen-etal-2017-reading} & TFIDF \cite{robertson2009probabilistic} & LSTM \cite{lstm} predicting the answer span from retrieved text \\ \hline
DPR \cite{karpukhin-etal-2020-dense} & BERT \cite{bert} based dual tower dense retriever & BERT \cite{bert} based model predicting answer span from retrieved text. \\\hline
RAG \cite{lewisRetrievalAugmentedGenerationKnowledgeIntensive} & DPR retriever & BART \cite{lewis-etal-2020-bart} based model generating answers using retrieved text. \\\hline
FiD \cite{izacardLeveragingPassageRetrieval2020} & DPR retriever & T5 \cite{2020t5} based model generating answers using retrieved text. \\ \hline
GAR \cite{mao2020generation} & BART \cite{lewis-etal-2020-bart} based model for query expansion; BM25 based retriever & DPR Reader (BERT based model predicting answer span from retrieved text.) \\ \hline
\end{tabularx}
\end{table}

\subsubsection{Retriever Methods for Factoid ODQA}
The goal of the Retriever in Retriever-Reader type ODQA systems is to search and find a small set of relevant abstracts which are then fed into the Reader. Therefore, the performance of the Retriever influences the performance of the entire system significantly. A variety of methods ranging from sparse lexical models to dense deep learning based models have been used for retrieving relevant passages. DrQA \cite{chen-etal-2017-reading} uses a TFIDF \cite{rajaraman_ullman_2011} based retrieval model to retrieve relevant Wikipedia passages. TFIDF and similar sparse lexical models like BM25 \cite{robertson2009probabilistic} have been used extensively for document retrieval in ODQA as well as other information retrieval tasks. First an inverted index of different words / word-tokens in the entire text corpus is created \cite{manning_raghavan_schutze_2008}. This produces a sparse high dimensional representation of the document. The query terms are used to match with the inverted index and a list of top scoring documents (scored by BM25 or TFIDF) is returned. 





These sparse retrieval models can also be modified to include higher order ngrams (bigrams or trigrams) along with the usual unigrams (individual word tokens), as they lead to better retrieval \cite{bigrams_retrieval}, albeit at the cost of reduced efficiency due to the increase in the number of features in the index. The sparse retrieval models only compare the exact matches, i.e. the lexical similarity among terms present in the documents and the query, and not their semantic relevance \cite{karpukhin-etal-2020-dense}. This is a major drawback which can be resolved using dense semantic encodings / embeddings produced by neural networks \cite{karpukhin-etal-2020-dense}.
\begin{figure}[h]
  \centering
  \label{fig:dual_bert}
  \includegraphics[width=\linewidth]{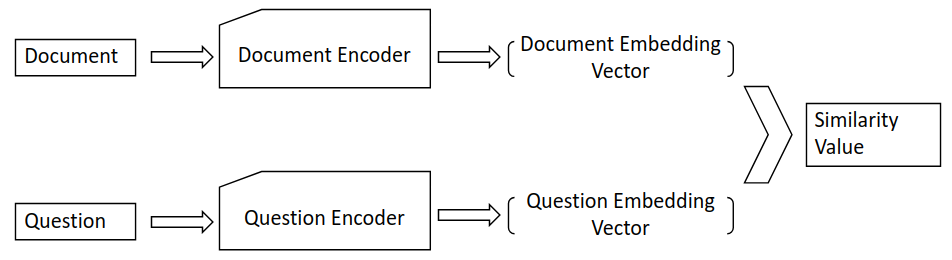}
  \caption{Dual Encoder or Two Tower architecture \cite{ChangYCYK20}. Query Encoder (also referred to as Query Tower) is used for encoding the question, Document Encoder (also referred to as Document Tower) is used to encode the documents. The encoders are usually BERT models. In ODQA retrieval models like DPR and ORQA \cite{karpukhin-etal-2020-dense, lee-etal-2018-ranking}, only the [CLS] output vector of the BERT model outputs embedding is used as the embedding vectors and the other token embeddings are ignored. The similarity value is usually the dot product among the normalized vectors.}
\end{figure}

Although neural networks have the potential to be more accurate than sparse lexical models for document retrieval, they are compute and time intensive if they compute the pairwise relevance of questions and documents together. For example, using cross-attention i.e. \textit{early fusion} based transformer architectures \cite{cross_attention_early} for calculating question-passage relevance for retrieval requires that during inference, every question-passage pairwise scores be calculated using the transformer model. This is prohibitively time consuming for large corpora with millions of passages. One methodology commonly used for finding relevant documents is to first retrieve a small set (usually tens to hundreds of abstracts) of relevant documents using a fast retriever method like BM25 and then rerank the set using a neural network \cite{r3, lee-etal-2018-ranking}. The top ranked documents (after the reranking step) can be used by the reader for predicting the answer. This was explored in a previous work \cite{lee-etal-2018-ranking} where it was shown that using an LSTM reranker on documents retrieved using BM25 model led to a significant increase in performance of the QA system. Another way of reranking was based on an end-to-end reranker-reader and reinforcement learning was described in R3 \cite{r3}. Most of the systems developed for ODQA uses a fixed number of retrieved or reranked documents to answer questions. However, it was empirically demonstrated that using a regression model to predict the number of passages to be used by the Reader can lead to better QA performance \cite{kratzwald-feuerriegel-2018-adaptive}.

Neural networks can also be used for the initial retrieval without sacrificing on the time needed. These models can be composed of dual BERT based encoders (also referred to as two-tower retrieval model) \cite{ChangYCYK20, Luan2021SparseDA} where one encoder is used for embedding the questions and the other one for the passages into a low dimensional continuous embedding space. The architecture has been schematically explained in figure 3. These models are trained so that the embeddings of the questions and passages are similar (in terms of dot product or cosine distance) in the embedding space if the question-passage pair are relevant, and dissimilar otherwise. The document embeddings are precomputed and stored, such that during inference, only the question embeddings need to be computed using the question encoder. The problem of retrieval can then be reduced to finding the K Nearest Neighbour document/passage embeddings to the question embeddings. Using specialized data structures like FAISS \cite{faiss} and Maximum Inner Product Search (MIPS), fast and efficient retrieval of relevant documents from a large collection is possible, making the method practically feasible, as was shown in the Dense Passage Retrieval (DPR) system \cite{karpukhin-etal-2020-dense}. DPR has been used as the retriever in several state of the art QA systems \cite{izacardDistillingKnowledgeReader2020, izacardLeveragingPassageRetrieval2020, lewisRetrievalAugmentedGenerationKnowledgeIntensive}. BERT models typically need a lot of labelled training data (question and relevant document pairs). In absence of such large datasets, the models can be pretrained using inverse close task \cite{lee-etal-2019-latent}, in which, instead of question and document pairs, sentences and its contexts (nearby passages) are used. 

Query modification, especially query expansion in document retrieval step has been shown to give rise to significant improvements in QA performance. The GAR \cite{mao2020generation} system uses a sequence to sequence generative transformer model, BART \cite{lewis-etal-2020-bart}, to generate query expansion terms. Using the expanded query in a simple BM25 model can cause significant improvement in the performance of the retriever. It was also shown that the GAR query expansion model used in conjugation with the DPR model can lead to even better performance.

\subsubsection{Reader Methods in ODQA}


The goal of readers in ODQA is to predict the factoid answers to a given question and a set of retrieved (and reranked) relevant documents. Since the retriever is rarely perfect, many of the documents fed to the reader models may not be relevant. Therefore the reader must be able to ignore irrelevant documents and form the answer using only the documents which can help in answering the question. This makes the task of a reader in ODQA or end-to-end QA tasks more complicated than that of a machine reading comprehension (MRC) task, where by design, only relevant documents are provided to the reader. Different types of systems have been used as the reader in ODQA task, out of which neural networks have been the dominant paradigm in recent years. 

\paragraph{Neural Models for Predicting Factoid Answers}
Currently neural reader models dominate the paradigm of open domain question answering. Initially, LSTM \cite{lstm} based answer span predictor model was used as the reader in the pioneering method named Dr. QA \cite{chen-etal-2017-reading, wang2018r}. However, gradually, with the suceess of large pretrained transformer based language models \cite{transformers} like BERT \cite{bert}, BART \cite{lewis-etal-2020-bart} and T5 \cite{2020t5}, they were quickly adapted to the task and showed remarkable progress. Currently these transformer models form the base of the state of the art reader models DPR\cite{karpukhin-etal-2020-dense}, RAG \cite{lewisRetrievalAugmentedGenerationKnowledgeIntensive} and FiD \cite{izacardLeveragingPassageRetrieval2020, izacardDistillingKnowledgeReader2020}.

One major reason for the success of transformer \cite{transformers} based pretrained models can be attributed to their apparent ability to store a large amount of knowledge, both linguistic as well as relational \cite{peters-etal-2018-dissecting, goldberg2019assessing, tenney2019learn, guuREALMRetrievalAugmentedLanguage2020, petroni-etal-2019-language}, which can be exploited in answering questions. These models can perform surprisingly well in QA tasks even without pretraining and, more interestingly, even without using any external information like retrieved documents \cite{petroni-etal-2019-language}. In the REALM system \cite{guuREALMRetrievalAugmentedLanguage2020}, it was further shown that adding a retrieval step significantly improves QA performance of transformer based models.

Based on the mechanism of answer prediction, the neural readers used in ODQA can be categorized into two types: 
\begin{enumerate}
    \item Extractive Models: They predict a span of the document provided as the answer. This can be done using an LSTM \cite{chen-etal-2017-reading, wang2018r} or a transformer like BERT \cite{karpukhin-etal-2020-dense}. The answers predicted from every retrieved document is combined using some heuristics to produce the final answers. The DPR system \cite{karpukhin-etal-2020-dense} comprises of a BERT \cite{bert} based extractive reader model, which takes a question and a set of passages as input, and is trained to predict a span in one of the passages as output.
    
    \item Generative Models: These models are essentially sequence to sequence models where, the questions and the documents are processed and an answer is sequentially generated \cite{lewisRetrievalAugmentedGenerationKnowledgeIntensive, izacardLeveragingPassageRetrieval2020}. In the RAG QA system \cite{lewisRetrievalAugmentedGenerationKnowledgeIntensive} the readers have been designed using BART \cite{lewis-etal-2020-bart}. In FiD \cite{izacardLeveragingPassageRetrieval2020}, an even more powerful T5 \cite{2020t5} based generative models are used.
\end{enumerate}

Currently, generative models \cite{izacardLeveragingPassageRetrieval2020, lewisRetrievalAugmentedGenerationKnowledgeIntensive} have outperformed the state of the art extractive models \cite{karpukhin-etal-2020-dense}. This can be attributed to two reasons. Firstly, generative models have the ability to combine the information present in different passages by combining the neural embeddings of every question-passage pair \cite{izacardLeveragingPassageRetrieval2020} before producing the answers. Secondly, whereas extractive models are strictly restricted to predicting spans in the retrieved documents as answers, generative models have the ability to produce any relevant text as the answer even if they are not present in the retrieved documents. This can be advantageous in certain cases where the ground truth answers do not correspond to any span of the document texts.

\paragraph{Information Retrieval Based Methods}

Before deep learning methods became popular, modified information retriever systems were commonly used for question answering. AskMSR \cite{brill-etal-2002-analysis} uses information retrieval primarily for finding sentences capable of answering a given question. It converts or reformulates a question such that the reformulations are likely to be present in the answer sentence. The different reformulations are weighted and used to retrieve relevant documents from the web. These documents are then mined and word n-grams are found. The n-grams are weighted using their frequency, the weights of the reformulation it corresponds to and also based on how well it matches the expected answer type of the question. Next, the n-grams are tiled (e.g. "A B C" and "B C D" can be tiled into "A B C D"), and the highest weighted n-grams are presented as the likely answers along with a probability score. Several other QA systems use information retrieval methods for machine comprehension. They are designed to predict an answer sentence (not a factoid answer) to a given question viz. DeepRead \cite{hirschman-etal-1999-deep}, QUARC \cite{quarc2000} and AQUAREAS \cite{ng-etal-2000-machine}. These models mainly used word count similarity or some variations to find the relevant answer sentences. 

\paragraph{Quasi Knowledge Graph Based Methods}
Another paradigm is to exploit the relationships among entities present in natural language text to construct a quasi knowledge graph and then processing it to extract the final answer \cite{lu2019quest, zhaoDelft2020}. 

QUEST \cite{lu2019quest} is an unsupervised method for complex factoid question answering using natural language text. The method first creates a noisy quasi knowledge graph using open information retrieval techniques from several relevant documents. Then it identifies the entities in the question and matches them to the entities in the quasi KG. A couple of interesting observations made by the authors is that the question entities usually appear in small groups in the quasi KG and the answers often lie in the paths connecting these question entity nodes. QUEST exploits these properties to identify the answer nodes. It finds several group steiner trees (GSTs) \cite{gst}. The GST reduces the search space for the answers significantly and the final answers are predicted using some heuristic criteria like answer type, relevance, presence of node term in question etc.

DELFT \cite{zhaoDelft2020} system constructs a quasi KG using text from Wikipedia, where entity mentions are nodes and the edges connecting entities are labelled by the sentence containing both the connected entities. When a question is given, the system identifies a relevant subgraph and prunes it using heuristics. Finally a graph neural network is used to find the representations of different nodes, which are further used by a multilayer perceptron to predict the answer nodes.

\subsection{Question Answering over Heterogenous Sources}
Question answering over structured knowledge bases and text have their own distinct advantages and disadvantages. Knowledge bases have a fixed schema and facts are expressed adhering to the schema which facilitates compositional reasoning \cite{das-etal-2017-question}. However, in order to be effective, the underlying knowledge bases used for KGQA need to have high coverage on the relations targeted by the questions. This is rarely seen as construction of knowledge bases require expert knowledge \cite{Paulheim2018HowMI,zhaoDelft2020}. Plus knowledge bases are notoriously incomplete \cite{dongKV2014, das-etal-2017-question}. Open Information Extraction (OpenIE) methods have been applied to extract relations from natural language text automatically without manual supervision, however, OpenIE methods favour precision over recall, and therefore are not able to solve the problem \cite{zhaoDelft2020}. On the other hand, QA over text is attractive as text may have potentially millions of facts hidden in them, which are not there in knowledge bases \cite{das-etal-2017-question}. However, QA over text is difficult primarily as text is unstructured, therefore not allowing compositional reasoning which is possible in structured knowledge bases \cite{das-etal-2017-question}. Additionally, there are several challenges associated with understanding of natural language text by computers, like ambiguity (same words, different meanings), synonymy (different words, same meaning) etc. 

In order to exploit the advantages and mitigate the disadvantages discussed above, a line of work focuses on exploiting both knowledge bases and text in question answering. One area where combination of KGs and text have been shown to be very useful is multihop question answering and multi-document question answering \cite{hotpotqa}. In spite of their advantages it must be remembered that combining KG and text is not trivial due to their structural non-uniformity \cite{das-etal-2017-question}.

A naive approach is to individually predict an answer using individual models for QA over text and QA over KGs, and then combine their predictions using some heuristics to produce the final answers \cite{Baudis2015ModelingOT}. This approach, (referred to as \textbf{late fusion}) have limited capacity to use evidence from different sources \cite{sun2018graftnet}. An alternative approach is to use \textbf{early fusion} in which the KG and the natural language text are fused into common data structures (graph or universal schema \cite{riedel2013univschema}) which are further processed to extract the answers \cite{das-etal-2017-question, sun2018graftnet, sun2019pullnet}

In UNISCHEMA \cite{das-etal-2017-question}, a universal schema \cite{riedel2013univschema} is constructed using entities and relations (triples) from a KG as well as related entities and texts where they are co-occurring \cite{das-etal-2017-question, riedel2013univschema}. This is used to construct a key-value memory \cite{miller-etal-2016-key}. For KG triples, the subject entity and the relation is used to construct the key, whereas for the textual facts, the subject entity and the text is used to construct the key. The neural representations of the keys and the question are computed using neural networks. These representations are further utilized by another iterative attention based neural network to predict the final answer entity. The entire model is trained end to end using question answer pairs. An alternate methodology of fusing KG facts and text into a subgraph was explored in Graftnet \cite{sun2018graftnet} and Pullnet \cite{sun2019pullnet}. In both these methodologies, entities present in the question is used to retrieve a quasi subgraph from the entire KG and retrieved Wikipedia text and graph neural network \cite{gnn} models are used to retrieve the answers.

\section{Biomedical Question Answering Overview}\label{BQAOV}
\subsection{Difference with general domain QA}
In spite of the progress in answering natural language questions in the general domain, the developed methods cannot be directly applied for answering questions in the domain of biomedicine \cite{zweigenbaum2003question}. This may be attributed to several reasons, the most notable among them being the following:
\begin{itemize}

    \item Language Specialization: The use of specialized jargon as well as their variations (e.g. abbreviations) make biomedical technical text very different from general text in English. Hence general domain QA systems which are not trained on the specialized language often fail.
    
    \item Limited information in Text Corpus: Most text based QA systems described in Section 2 has been designed to work with Wikipedia text and cannot directly be used on other corpora viz. PubMed without suitable retraining. Although an excellent source of information in a broad variety of fields, Wikipedia lacks depth and might not contain up-to-date information in the fast evolving domain of biomedicine. 
    
    \item Limited information in Knowledge Bases: One reason for the success of general domain QA systems can be attributed to the development and exploitation of large knowledge bases like DBPedia \cite{dbpedia}, Freebase \cite{bollacker2007freebase}, Yago \cite{fabian2007yago}, Wikidata \cite{vrandevcic2014wikidata} etc. However these KBs primarily consists of general domain knowledge and do not contain fine grained knowledge in the domain of biomedicine. Therefore these cannot be easily applied to answer biomedical questions.
    
\end{itemize}

\subsection{Biomedical Question Answering Benchmark Datasets} \label{BQAD}
Several biomedical question answering (BQA) datasets have been described in literature. When compared to their general domain counterparts, biomedical QA benchmark datasets are much smaller in number of data points (question answer pairs). This is primarily because creation of biomedical QA datasets need expert curation, making the process very expensive. Therefore, the largest BQA benchmark datasets are an order of magnitude smaller than general domain QA benchmarks. Similar to what we had discussed for the general domain, biomedical QA systems can use either unstructured text (biomedical scientific articles) or structured knowledge bases for answering questions. Therefore, a primary way of classifying biomedical benchmark datasets would be on the basis of the source of knowledge used:

\begin{itemize}
    \item \textbf{Text Based Biomedical QA Benchmarks}
    The focus of majority of the biomedical QA systems has been on using PubMed articles as the source of knowledge. The overall summary of benchmarks developed for training and evaluation of those systems are listed in table \ref{tab:bqadataset}
    
    \item \textbf{Knowledge Based Biomedical QA Benchmarks}
    Although the predominant paradigm for BQA is using scientific literature text, few benchmarks of limited scope and size have also been developed for biomedical QA over linked data (knowledge graphs). These have been listed in table \ref{tab:bqadatasetkg}.
\end{itemize}

Apart from the source of knowledge used for biomedical QA, based on their characteristics viz. task, domain and construction, text based biomedical QA benchmark datasets, specifically the text based BQA benchmark datasets, can be classified in many ways.

\subsubsection{Classification by Task}

\begin{itemize}
    
    \item \textbf{Retrieval}
    
    The task of retrieval, as the name suggests, involves finding a set of documents from the entire corpus (e.g. PubMed), that contain relevant information for answering a given question, for e.g., the BioASQ Task B Phase B \cite{bioasq2015}.
    
    \item \textbf{Answer Matching}
    
    For this, a large collection of possible long answers are collected, and the main objective of the task is to identify the correct answer to a given question \cite{cmedqa1, cmedqa2}.
    
    \item \textbf{Machine Reading Comprehension (MRC)}
    
    A question and one or a few relevant passages containing information for answering a question is given, and the primary task is to predict the correct answer \cite{medqamixed, jin2019pubmedqa, bioasq2015}.
    
\end{itemize}
\subsubsection{Classification by Answer Type}
Based on the type of answer, the benchmark datasets can be classified into the following types:

\begin{itemize}
    \item \textbf{Factoid}
    The answers are short pieces of text comprising of a few words \cite{bioasq2015}.

    \item \textbf{List}
    The answers are a list or unordered set of several short factoid answers \cite{bioasq2015}.
    
    \item \textbf{Yes/No}
    The answer is a binary "Yes" or "No". \cite{bioasq2015, jin2019pubmedqa}
    
    \item \textbf{Summary}
    The answers are short passages. \cite{bioasq2015}
    
    \item \textbf{Multiple Choice Question (MCQ)}
    A few answer options are provided out of which one or more are correct. \cite{vilares2019head, li2020nlpec}.

\end{itemize}

\subsubsection{Classification by Sub-Domain}
Within the domain of biomedicine, question answering datasets can be classified into the following subdomains depending upon its intended audience and type \cite{biomedicalqareview}:

\begin{itemize}
    \item \textbf{Scientific}: The questions are typically asked by biomedical scientists and clinicians. The questions are scientific in nature, and the answers are typically found in scientific literature \cite{bioasq2015}.
    
    \item \textbf{Consumer}: These questions are typically asked by consumers or non-medical audience who are mostly interested in knowing more about their clinical conditions, treatments etc. \cite{cmedqa1, cmedqa2}.
    
    \item \textbf{Medical Examination}: These datasets are constructed from questions asked in medical certification or other associated exams. These comprise of clinical questions which needs both knowledge and logical reasoning to answer \cite{vilares2019head}.
\end{itemize}

\begin{table}[t]
\center
  \caption{Biomedical Question Answering using Structured Data}
  \label{tab:bqadatasetkg}
    \centering
    \resizebox{\textwidth}{!}{%
    \begin{tabular}{lllllll}
    \hline
    \textbf{Name of Dataset} & \textbf{Task} & \textbf{Data Sources} & \textbf{\begin{tabular}[c]{@{}l@{}}Formal \\ Queries\end{tabular}} & \textbf{Domain} & \textbf{\begin{tabular}[c]{@{}l@{}}Question \\ Language\end{tabular}} & \textbf{\begin{tabular}[c]{@{}l@{}}Number of \\ QA Pairs\end{tabular}} \\ \hline
    QALD4 Task 2 \cite{qald4} & \begin{tabular}[c]{@{}l@{}}QA Over \\ KG\end{tabular} & \begin{tabular}[c]{@{}l@{}}KG (Linked Data) using 3 databases viz. \\ Diseasome \cite{diseasome}, DrugBank \cite{drugbank2018} and SIDER \cite{sider}.\end{tabular} & Yes. In SPARQL & Scientific & English & 50 \\ \hline
    Bioinformatics \cite{biosoda} & \begin{tabular}[c]{@{}l@{}}QA Over \\ KG\end{tabular} & \begin{tabular}[c]{@{}l@{}}KG (Linked Data) using 2 databases viz. \\ BGee \cite{bgee}and OMA \cite{oma} \end{tabular} & Yes. In SPARQL & Scientific & English & 30 \\ \hline
\end{tabular}%
}
\end{table}

\begin{table}[]
\centering
\caption{Different Benchmark Datasets for Biomedical Question Answering. Here several benchmarks for BQA and associated tasks (other than end-to-end question answering) have also been included for completeness. Other benchmark datasets which are not in English have been excluded.}
\label{tab:bqadataset}
\resizebox{\textwidth}{!}{%
\begin{tabular}{llllll}
\hline
\textbf{Dataset} & \textbf{Task} &  \textbf{\begin{tabular}[c]{@{}l@{}}Type of \\ Answers\end{tabular}} & \textbf{Domain} & \textbf{Language} & \textbf{\begin{tabular}[c]{@{}l@{}}Number of \\ Questions\\  in Dataset\end{tabular}} \\ \hline
BioASQ Task B Phase A & Retrieval & - & Scientific & English & 3742 \\ \hline
BioASQ Task B Phase B & MRC & \begin{tabular}[c]{@{}l@{}}Yes/No + \\ Factoid + \\ List + \\ Summary\end{tabular} & Scientific & English & 3742 \\ \hline
MedQA \cite{medqamixed} & MRC & MCQ & \begin{tabular}[c]{@{}l@{}}Medical \\ Examination\end{tabular} & \begin{tabular}[c]{@{}l@{}}Chinese \\ and English\end{tabular} & ~12000 \\ \hline
HEADQA \cite{vilares2019head} & MRC & MCQ & \begin{tabular}[c]{@{}l@{}}Medical \\ Examination\end{tabular} & \begin{tabular}[c]{@{}l@{}}Spanish \\ and English\end{tabular} & ~6700 \\ \hline
PubMedQA \cite{jin2019pubmedqa} & MRC & Yes/No & Scientific & English & ~260000 \\ \hline
BioMRC \cite{pappas2020biomrc} & Cloze Task & - & Scientific & English & ~812000 \\ \hline
Biomed-Cloze \cite{biomedcloze} & Cloze Task & - & Scientific & English & ~1 Million \\ \hline
\end{tabular}
}
\end{table}

\subsection{Some Important BQA Datasets}

\subsubsection{BioASQ Task B}
BioASQ Task B \cite{bioasq2015} is arguably the most popular question answering benchmark dataset in the domain of biomedicine. It is a challenge (competition) consisting of two subtasks which are referred to as Phases. Phase 1 is essentially document retrieval i.e. the task of finding article abstracts from the entire PubMed corpus, to a given question, such that the question can be answered by information present in the retrieved articles. Phase 2 on the other hand deals with machine reading comprehension, which pertains to answering the question when several relevant abstracts are provided. The BioASQ \cite{bioasq2015} dataset currently comprises of more than 4 thousand expert curated training questions, with 500 new questions added every year.  Along with each question, relevant scientific abstracts which contain the answer and the answer are also provided. For some questions, the URI of the concepts in biomedical KBs and relevant KB triples are also provided. The dataset contains roughly equal numbers of questions of the following four answer types factoid, list, yes/no and summary.

The principal reasons for the popularity of the BioASQ challenge and dataset is because of its manual (expert) construction and large size (compared to other manually annotated biomedical QA datasets). However, the BioASQ dataset \cite{bioasq2015} is not free from errors. 

Although each question is marked by annotators as belonging to one of the types, there are several errors. For e.g. consider the question from the challenge dataset:

    "What is the combined effect of Nfat and miR-25?" 
    
This is marked as a factoid question. However, the annotated answer is:

    "Re-express the basic helix-loop-helix (bHLH) transcription factor dHAND (also known as Hand2) in the diseased human and mouse myocardium.".
    
The above answer cannot be considered not a factoid answer, since factoid answers in general can be represented in a few words or less than 50 bytes \cite{soricut-brill-2004-automatic}. Such examples are not very hard to come by in the dataset. Additionally, the method for evaluating the factoid or list answers are based on exact string matching which does not take into account that biomedical entity names contain several variations (viz. abbreviations, synonyms etc.). Therefore, correctly predicted answers may be marked as incorrect by the evaluation method. Moreover, along with every question, a list of abstracts which can answer the question is provided, and these serves to evaluate the performance of the retrieval systems. However, the list of abstracts which can answer each question is incomplete, and several questions can be answered by several abstracts which are not in the ground truth list. This leads to correctly retrieved articles being marked as incorrect. Finally, the concepts and triples provided along with some questions are not very reliable, which might explain why most research groups tend to focus on the development of text based QA techniques, rather than trying to utilize structured knowledge bases. Finally, since the QA pipeline in the BioASQ challenge \cite{bioasq2015} has been divided into two subtasks which are scored and evauated in isolation, several state of the art methods which focus on one of these two tasks in isolation have been developed. End-to-end question answering, which has more practical use, has remained relatively neglected. Despite these limitations, BioASQ remains the most popular benchmark for training and testing biomedical QA systems.

\subsubsection{PubMedQA}
PubMedQA \cite{jin2019pubmedqa} is a question answering dataset over biomedical (PubMed) abstracts. It is a machine reading comprehension dataset where the systems have to understand the abstract to answer the question with a "yes"/"no"/"maybe". It contains nearly 1 thousand manually curated and more than 200 thousand artificially generated QA pairs.

\subsubsection{Cloze Type QA datasets}
Since manually annotating question answer pairs in the domain of biomedicine is expensive,  several "cloze type" QA datasets have been developed using automated methods\cite{pappas2018bioread, pappas2020biomrc, dhingra2018simple}. In brief, to develop a QA pair, a passage consisting of multiple sentences is taken and divided into a large (several sentences) and a small (one sentence) passages. A named entity present in the smaller sentence is replaced by a placeholder. The task of the ML model is to predict the replaced entity using the two texts. The larger and the smaller text can be thought to be similar to a context and a question respectively. Such datasets have been used for pretraining biomedical QA models \cite{dhingra2018simple}, although they cannot be considered as QA benchmarks.

\subsubsection{Re-purposing Medical Examination Questions}
Another strategy of creating biomedical question answer pairs is reusing questions from examinations in medicine and related fields. A few such QA datasets have been constructed in the last few years \cite{vilares2019head, li2020nlpec}. A majority of such exams are composed of multiple choice questions and may or may not have a context to extract the answer from. However, it should be remembered that MCQ question answering is inherently different from the more general open domain paradigm. Hence systems trained on MCQ questions cannot be directly applied to answer open domain questions and will require retraining or fine tuning.

\subsubsection{Other Datasets}

Several other datasets focusing on one or a few components of a question answering system. For e.g. the MEDIQA datasets \cite{mediqa1} are focused towards several tasks such as natural language inference, recognizing question entailment, consumer health question summarization, multi-answer summarization, radiology report summarization, and their applications in medical Question Answering (QA). Although some of these datasets can help in improving biomedical and clinical question answering systems \cite{biobertbioasq8b}, they cannot be used in isolation for developing end-to-end open domain QA methods. 

Some popular biomedical question answering datasets along with their characteristics are listed in Table \ref{tab:bqadataset} and Table \ref{tab:bqadatasetkg}. For a more comprehensive review of different associated tasks such as question entailment, answer summarization and visual question answering, refer to \cite{bqareview}.

\subsection{Methodologies}
Similar to general domain question answering, biomedical question answering can be divided into the following approaches:
\begin{enumerate}

    \item Biomedical Question Answering using Structured Knowledge
    \item Biomedical Question Answering using Text

\end{enumerate}

These will be discussed in detail in the following sections. For simplicity, we will include open domain (end to end) QA, machine comprehension as well as information retrieval tasks within BQA using Text. Other modalities of biomedical question answering like visual question answering using medical images which have gained popularity recently will be excluded those from this review.

\section{Biomedical Question Answering using Structured Knowledge} \label{BQAKB}
A large majority of the BQA systems developed in recent years belong to the paradgm of QA using text, which can be attributed partly due to the availability and popularity of the BioASQ \cite{bioasq2015} challenge benchmark datasets. Only a few BQA systems using structured knowledge have also been described \cite{biosoda, lodqa, pomelo2, gfmed}. Using structured knowledge bases has certain advantages in QA. Structured knowledge bases are easily understood by computers in comparison to unstructured natural language text. Several QA systems using structured knowledge such as knowledge graphs as the source of information have been successfully developed in the general domain (described in Section 2). Also knowledge graphs are better suited to explaining the generated answers, which is another important aspect, especially for BQA. Additionally, there has been a proliferation of curated databases in the field of biomedicine, each specializing in a particular domain of biology and medicine, viz. UniProt (knowledge about proteins) \cite{uniprot2021}, DrugBank (information about drugs) \cite{drugbank2018} etc. These can be potentially used by biomedical QA systems as a source of information. However, in order to fully exploit the potential of the knowledge present in these databases, the information from different databases should be unified into a interlinked or connected knowledge base. Recently, a variety of graph analytic techniques have been applied to biomedical knowledge graphs for a variety of tasks such as prediction of drug adverse effects \cite{sideeffectsdrug}, drug targets \cite{deeppurpose}, diseases from symptoms \cite{NEURIPS2020_5bca8566} and several others \cite{li2021representation}. Also, several techniques for querying over incomplete knowledge graphs have also been developed \cite{NEURIPS2018_ef50c335, lin-etal-2018-multi-hop, guo-etal-2016-jointly, xiong-etal-2017-deeppath, guu-etal-2015-traversing, yang2014embedding, das2017chains}. Moreover QA systems developed recently have the ability to reason and make inferences over KGs \cite{lego, qagnn}. Having similar capabilities in biomedical QA systems has the potential push the boundaries of healthcare. In spite of all these advantages, BQA using KGs has been relatively unexplored, primarily because the dearth of large scale benchmark datasets and the lack of a unified knowledge base or knowledge graph that assimilates the information present in the disparate databases.

\begin{table}[]
\centering
\caption{List of a small subset of biomedical databases and ontologies which may be potentially useful for BQA. All of these databases are curated using a mix of manual and computational methods. The primary entity type of each of the databases are in \textbf{bold}.}
\label{tab:dbs}

\resizebox{\textwidth}{!}{%

\begin{tabular}{llll}
\hline
\textbf{Database} & \textbf{Entity Types} & \textbf{\begin{tabular}[c]{@{}l@{}}Relations and other \\ Information Present\end{tabular}} & \textbf{Type of Database} \\
\hline
UniProtKB \cite{uniprot2021} & \begin{tabular}[c]{@{}l@{}}\textbf{Proteins}, Diseases, \\ Genes, Species\end{tabular} & \begin{tabular}[c]{@{}l@{}}Protein-Disease, \\ Protein-Gene, \\ Protein-Function\end{tabular} & Tabular and Textual Database \\ \hline
DrugBank \cite{drugbank2018} & \begin{tabular}[c]{@{}l@{}}\textbf{Drugs}, Protein (Target), \\ Disease, Commercial Product \\ Names, Side Effects, \\ Adverse Effects\end{tabular} & \begin{tabular}[c]{@{}l@{}}Drug-Disease, \\ Drug-Target, \\ Drug Classification\end{tabular} & Tabular and Textual Database \\ \hline
CHEMBL \cite{chembl} & \begin{tabular}[c]{@{}l@{}}\textbf{Drugs}, Protein (Target), \\ Disease, Commercial \\ Product Names etc.\end{tabular} & \begin{tabular}[c]{@{}l@{}}Drug-Disease, \\ Drug-Target, \\ Similar Chemicals etc.\end{tabular} & Tabular / Graph Database \\ \hline
\begin{tabular}[c]{@{}l@{}}GO \cite{go2021} and \\ associated databases \\ like QuickGO \cite{quickgo2019}\end{tabular} & \begin{tabular}[c]{@{}l@{}}\textbf{Gene Product Function}\\ \textbf{and Location}, \\ Anatomy, Biochemicals, \\ Cells etc.\end{tabular} & \begin{tabular}[c]{@{}l@{}}Gene/Protein Function,\\ Protein-Location, \\ Biochemical Reactions, etc.\end{tabular} & Ontology \\ \hline
Disease Ontology \cite{doid} & \textbf{Diseases}, Symptoms & \begin{tabular}[c]{@{}l@{}}Disease Classification, \\ Disease-Symptoms\end{tabular} & Ontology \\ \hline
OMIM \cite{omim} & \textbf{Human Genes and Diseases} & \begin{tabular}[c]{@{}l@{}}Gene-Disease, \\ Disease Inheritance\end{tabular} & Tabular and Textual Database \\ \hline
DisGeNet \cite{disgenet2019} & \begin{tabular}[c]{@{}l@{}}\textbf{Human Genes and Diseases},\\  Mutations\end{tabular} & \begin{tabular}[c]{@{}l@{}}Gene-Disease, \\ Mutations-Disease\end{tabular} & Tabular and Textual Database \\ \hline
MESH \cite{mesh} & \begin{tabular}[c]{@{}l@{}}Most if not all types \\ of biomedical entities\end{tabular} & \begin{tabular}[c]{@{}l@{}}Hierarchical Classification of \\ Medical Terms and their \\ Descriptions\end{tabular} & Ontology \\ \hline
SNOMEDCT \cite{snomed} & \begin{tabular}[c]{@{}l@{}}Most if not all types \\ of biomedical entities\end{tabular} & \begin{tabular}[c]{@{}l@{}}Hierarchical Classification of \\ Medical Terms and their \\ Descriptions, Disease-Causes, \\ Disease-Treatments etc.\end{tabular} & Ontology \\ \hline
UMLS \cite{umls2004} & \begin{tabular}[c]{@{}l@{}}Most if not all types of \\ biomedical entities\end{tabular} & \begin{tabular}[c]{@{}l@{}}Metathesaurus i.e mappings \\ of entities in different databases.\end{tabular} & Thesaurus \\ \hline
NCBI Taxonomy \cite{ncbi_taxonomy2020} & \textbf{Organism Names} & \begin{tabular}[c]{@{}l@{}}Hierarchical Classification \\ of Organisms and their \\ Different Names\end{tabular} & Taxonomy \\ \hline
HMBD \cite{hmdb} & \begin{tabular}[c]{@{}l@{}}\textbf{Human Metabolites}, \\ Diseases, Anatomy etc.\end{tabular} & \begin{tabular}[c]{@{}l@{}}Metabolite-Disease, \\ Metabolite-Anatomy\end{tabular} & Tabular and Textual Database \\
\hline
\end{tabular}%
}
\end{table}

\subsection{Biomedical Structured Knowledge Resources} \label{BKB}
In the domain of biology, there are several databases which contain information about different biological entities and their relations. The properties of a few important databases are listed in Table \ref{tab:dbs}. The number of such databases have grown rapidly in the last twenty years. We have classified different types of databases and knowledge bases commonly encountered in the domain of biomedicine into these 4 categories, expanding upon the classification found in \cite{kurdi2020generation}:
\begin{enumerate}
    \item Tabular or Textual Databases: These databases contain the relationships among narrow types of biomedical entities in the form of tables or a mixture of tabular and text data. For e.g., DisGenet \cite{disgenet2019} contains the associations between genes, mutations and diseases, QuickGO (GO \cite{go2021} Annotations) contains the functions and locations of gene products like proteins \cite{quickgo2019}, UniProtKB \cite{uniprot2021} consists of a plethora of information about proteins, etc.
    
    \item Taxonomy describes a hierarchical relationships among different entities mainly consisting of hypernyms and hyponyms, for e.g., "Tiger is\_a Mammal", "Human is\_a Primate" etc. The most commonly used taxonomical knowledge base in biomedicine are the taxonomic classification of life forms like NCBI Taxonomy \cite{ncbi_taxonomy2020}. 
    
    \item Thesauri (singular: Thesaurus) are similar to taxonomies, containing hierarchical relations, along with other predefined relations such as synonymy and associations, for e.g. "Rats associated\_with Bubonic Plague", "Black Rat same\_as Rattus rattus". The best known thesaurus in the domain of biomedicine is UMLS (Unified Medical Language System) \cite{umls2004}.
    
    \item Ontologies describe entities, instances of entities as well as relations among them. They comprise of a much broader set of relations than taxonomies and thesauri. Ontologies can express axioms and restrictions. On one hand this maximizes the of expressivity of ontologies, on the other hand, they are hard to create, maintain and handle. Examples of well known ontologies in the domain of biology include Gene Ontology \cite{go2021} and Disease Ontology \cite{doid}.

\end{enumerate}

A majority of these databases are manually annotated and focus on molecular biological functions, interactions and relations such as Uniprot \cite{uniprot2021}, DrugBank \cite{drugbank2018}, NCBI Taxonomy \cite{ncbi_taxonomy2020}, DisGenet \cite{disgenet2019} to name a few. Alongside such databases of relations among different biological entities and their functions, several ontologies have also been developed. The most popular among them being the Gene Ontology \cite{go2021} which contains structured knowledge about gene functions (GO terms) and also the relationships between them. Other ontologies include Disease Ontology \cite{doid}. Usually domain agnostic knowledge graphs are not helpful in BQA, in certain cases, KGs such as Wikidata can answer biomedical questions. Most of specialized databases are typically very narrow in scope, therefore their use is restricted to narrow applications \cite{bio2rdf2008, bioKG2019}. Therefore a number of research efforts has been directed towards the fusion of the disparate database into a useful resource spanning across several biomedical entity classes and relations \cite{kbbiods2020, bio2rdf2008, bioKG2019, hetionet2017}.

Although a lot of effort has been directed into the manual extraction of structured knowledge about molecular biological entities, less focus has been provided into manually annotating medical entities and their relationships. For e.g., a comprehensive manually curated database of diagnostic tests and their corresponding diseases cannot be found. Therefore, information extraction tools have been developed and utilized. SemRep \cite{semrep2020}, an information extraction tool has been used for extracting relationships among biomedical entities in biomedical literature texts and the extracted information has been used to construct the SemMedDB (Semantic Medline DB) \cite{semmeddb2012}. Another database of note is the UMLS (Unified Medical Language System) which serves as a metathesaurus in the field of biomedicine, and provide links to several other commonly used biomedical knowledge bases \cite{umls2004}. Some important biomedical databases and ontologies are described in brief. A more comprehensive list of databases which can be used to answer several types of commonly asked biomedical questions is listed in Table \ref{tab:dbs}

\begin{itemize}
    \item Uniprot \cite{uniprot2021} is a comprehensive database of information about proteins. They consist of the protein sequence, function, location along with other information. It consists of manually annotated information from scientific literature (SwissProt) as well as computationally predicted information (TremBL).
    
    \item Gene Ontology \cite{go2021} 
    consists of three detailed ontologies of terms describing the molecular functions of genes and proteins, the biological processes they perform and their subcellular locations. There are several databases built using GO, for e.g. QuickGO \cite{quickgo2019} contains GO annotations of genes and gene products, and is linked to Uniprot \cite{uniprot2021}.
    
    \item DrugBank \cite{drugbank2018} is a knowledge base of information about different drugs, including their physical and chemical properties, pharmacodynamics, commercial names, protein targets, pathways, enzymes and transporters involved with the drug etc.
    
    \item UMLS \cite{umls2004} consists of a metathesaurus as well as a semantic network that brings together over 2 million names for around 90 thousand concepts from more than 60 families of biomedical and health vocabularies, as well as 12 million relations among these concepts. Databases like SemMedDB \cite{semmeddb2012} are constructed using the UMLS entities and relations.
    
    \item Semantic Medline Database (SemMedDB) \cite{semmeddb2012} is an automatically constructed knowledge base (using SemRep \cite{semrep1, semrep2, semrep2020} for information extraction) from biomedical scientific literature. It is built using UMLS, i.e. the entities and relations correspond to ones present in UMLS. However, since the database is automatically annotated, its accuracy is low \cite{kbbiods2020}

\end{itemize}

Apart from the ones discussed, there are several other useful knowledge bases which can be used for biomedical question answering. Readers are encouraged to refer to \cite{li2021representation, kbbiods2020, bioKG2019} to get a more comprehensive list of such knowledge bases and their applications. 

\subsection{Methods}

SemBT \cite{sembt2015} describes a system for answering semi structured text queries using a database of relations among biomedical entities. The relations were extracted from the entire Medline/PubMed corpus using the SemRep \cite{semrep2020, semrep1} relation extraction tool and stored in a SQL database. The major advantage of the method is that the underlying database, and consequently the question answering system has broad coverage and contains entities and relations of several types of biomedical classes. The major disadvantage of the system is that it is capable of answering only simple semi structured queries and cannot interpret natural language queries. Although it is easier than learning to write structured queries in SQL, users need to be familiar with the semi structured query language method to use the system. Another weakness of the system is that it is able to answer only simple queries or one hop queries, while practical applications need provision for multihop questions as well. Finally the method depends on a automatically constructed database of relations i.e. SemRep, which has been shown to extract inaccurate relations \cite{semrep2}. So user discretion and verification is needed before using the produced answers can be used in any decision making process.

MEANS \cite{abacha2015means} also involves question answering over a structured knowledge base focusing on the medical domain. The knowledge base is constructed using customized automatic information extraction techniques, employing both machine learning as well as rule based methods. However, the MESA ontology on which the knowledge base is based is restricted and focused towards specific medical questions. The method for QA involved processing natural language questions into a SPARQL query using the MESA ontology, exploiting rule based and machine learning based techniques. The MEANS system also consists of methods of query relaxation (to reduce the restrictions of the SPARQL queries) and answer reranking for producing the final answer. The MEANS system has several desirable qualities like ability to process natural language queries into structured formal queries and the ability to answer multihop complex questions. However, the domain of the system is restricted to only a few types of medical entities and relations, for e.g. treatment, test, disease, etc. Therefore, MEANS cannot be used to answer biological questions, or even deeper medical questions involving concepts not described in the MESA ontology.

LODQA (Linked Open Data Question Answering) \cite{lodqa} is a system for generating structured queries in SPARQL from natural language questions using the SNOMEDCT ontology. The methodology can be divided into the following steps. First the natural language questions are processed to identify the noun chunks representing the subject and the target. The subject and object terms are used to search and find the corresponding URIs from biomedical ontologies. Any entity belonging to the target type having a relation with the subject can be the answer. However, there might be several possible answers using the above step and although this will increase sensitivity, therefore there is a need to filter the relations and reduce the possible answer types to increase specificity. The shortest dependency path between the subject and target in the question surface form can also be identified and used to identify and filter the relations among the subject and target, thereby increasing specificity of the method. Although several interesting ideas are mentioned in the paper the method has not been evaluated. Neither has there been any follow up work expanding upon this method.

Several KGQA systems have been developed for the QALD biomedical QA task \cite{qald4}. GFMed \cite{gfmed} is a methodology for QA using the combined KG using 3 linked databases viz. DrugBank \cite{drugbank2018}, SIDER \cite{sider} and Diseasome \cite{diseasome} introduced in the QALD Task 2 \cite{qald4}. Thes method depends on using manually defined grammatical rules for the conversion of question sentences in English to a corresponding SPARQL query that is executed over the combined KG to give the answer. Additionally the paper also describes a general pipeline for converting questions asked in other languages to SPARQL queries, and demonstrate the feature using Romanian language to ask questions. POMELO \cite{pomelo2} is another similar system which is aimed at answering natural language questions using the aforementioned unified KG by converting the question into a corresponding SPARQL query. The major drawback of the above mentioned systems, viz. GFMed \cite{gfmed} and POMELO \cite{pomelo2} is its use of handcrafted grammatical rules which cannot generalize to every possible practical scenario. This can lead to problems in practical settings, where much varied vocabulary may be used and the unified KG might not be limited to the 3 aforementioned databases. 

Bio-SODA \cite{biosoda} is quite similar to POMELO \cite{pomelo2} and GFMed \cite{gfmed}, in that it can answer questions over the unified KG from QALD Task 2 \cite{qald4} by translating the questions into corresponding SPARQL queries. However, it has the added ability to potentially adapt to any given knowledge graph. The system is not trained using data, neither does it depend on complex grammatical rules. Rather, it finds the answer using the terms present in the question and graph analytics algorithms. Although in theory this method is agnostic of the KG used, practically, it would suffer if the KG schema becomes too broad.

\begin{table}[]
\centering
\caption{Some methods for answering biomedical questions using structured knowledge either extracted using automated methods or manually curated.}
\resizebox{\textwidth}{!}{%
\begin{tabular}{lllll}
\hline
Method & Knowledge Base & Question & Question Complexity & Answer Types \\ \hline
SemBT \cite{sembt2015} & \begin{tabular}[c]{@{}l@{}}SQL database constructed using SemRep \cite{semrep2} \\ information extraction tool\end{tabular} & \begin{tabular}[c]{@{}l@{}}Natural Language\\ but similar\\ to Formal Queries\end{tabular} & mostly simple & Factoid, List \\ \hline
MEANS \cite{abacha2015means}& \begin{tabular}[c]{@{}l@{}}KG constructed using custom automatic \\ information extraction\end{tabular} & Natural Language & simple + complex & Factoid, List \\ \hline
LODQA \cite{lodqa} & SNOMEDCT \cite{snomed} & Natural Language & simple + complex & Factoid, List \\ \hline
GFMed \cite{gfmed} & \begin{tabular}[c]{@{}l@{}}QALD Linked Data KG (Diseasome \cite{diseasome}, \\ DrugBank \cite{drugbank2018}, SIDER \cite{sider})\end{tabular} & Natural Language & simple + complex & Factoid, List \\ \hline
POMELO \cite{pomelo2} & \begin{tabular}[c]{@{}l@{}}QALD Linked Data KG (Diseasome, \\ DrugBank, SIDER)\end{tabular} & Natural Language & simple + complex & Factoid, List \\ \hline
Bio-SODA \cite{biosoda} & \begin{tabular}[c]{@{}l@{}}KG Agnostic (Claim). Tested on QALD Linked \\ Data KG and Bioinformatics KG\end{tabular} & Natural Language & simple + complex & Factoid, List \\ \hline
\end{tabular}%
}
\end{table}

\section{Biomedical Question Answering using Text} \label{BQAT}
Natural language text is the most commonly used resource used by automated biomedical question answering systems. The most commonly used text resource used for BQA is the PubMed corpus. Since the PubMed corpus is huge, containing nearly 29 million abstracts, text based BQA systems, like general domain ODQA systems, usually consist of two parts: a retriever for finding a subset of abstracts which are relevant for answering a given question, and a reader for predicting the answer given the question and relevant abstracts. Different types of rule based and machine learning based systems have been described in the past. However, due to their success and potency in different types of NLP tasks, transformer \cite{transformers} based pretrained language models like BERT \cite{bert}, ElMo \cite{elmo} etc. have also been adapted to biomedical question answering in recent years, and currently the most commonly used paradigm for developing BQA systems.

In general ODQA setting, different text resources have been used as sources of information, like the entire web, wikipedia, books etc. Wikipedia is currently used as the text source of choice in most QA systems, because of the breadth of information contained in it and its reliability. However, Wikipedia lacks the depth as well as the reliability and may not always be up to date, therefore other resources are required for biomedical QA systems. PubMed, a bibliographic database consisting of nearly 30 million journal articles along with their abstracts (for the majority of the abstracts), contain most accurate and fresh information in the domain of biomedicine. Biomedical literature abstracts found in PubMed can serve as the source of knowledge for QA and have been used in several biomedical QA benchmark tasks like BioASQ \cite{bioasq2015}, PubMedQA \cite{jin2019pubmedqa} etc.  Since the corpus of text in PubMed is enormous, similar to general domain ODQA using text, retriever-reader architectures are used for BQA as well.
    


\subsection{Methods for Retrieval and Answering Summary Type Questions}

Document and passage retrieval is the first step for any open domain or end-to-end QA tasks. Additionally there are several methods which aim to retrieve relevant documents, passages as well as individual sentences or snippets and produce them as the answers \cite{semrepsum, cqa, essie, bioanswerfinder19}. In this section we focus on methods for retrieving documents as well as snippets within documents with the goal of answering natural language questions. For continuity, we also include methods for producing summary type answers to questions in this section. Based on the techniques used, the systems can be classified into:
\begin{enumerate}
    \item Classical {Lexical or Rule Based Methods}
    \item Neural Methods
\end{enumerate}

\begin{table}[]
\centering
\caption{Methods for document and snippet retrieval as well as generating summary type answers to given questions.}
\resizebox{\textwidth}{!}{%
\begin{tabular}{llll}
\hline
\textbf{Method} & \textbf{Task} & \textbf{Method Classification} & \textbf{Details} \\ \hline
BM25 \cite{robertson2009probabilistic} & Document Retrieval & Classical method & Similar to open \\ \hline
Sarrouti \cite{sarrouti2016generic} & Document Retrieval & Classical method & \begin{tabular}[c]{@{}l@{}}Uses UMLS concepts and \\ UMLS semantic network to\\ calculate semantic similarity \\ among question and document \\ title.\end{tabular} \\ \hline
SemRep Summaries \cite{semrepsum} & Document Retrieval & Classical method & \begin{tabular}[c]{@{}l@{}}Exploit relationships extracted\\ using SemRep tool to enhance\\ document search.\end{tabular} \\ \hline
Kumar \cite{naresh-kumar-etal-2018-ontology} & \begin{tabular}[c]{@{}l@{}}Summary Answer\\ Generator\end{tabular} & Classical method & \begin{tabular}[c]{@{}l@{}}Exploits relationships extracted\\ using custom information\\ extraction tool to retrieve \\ relevant text snippets\end{tabular} \\ \hline
Ask-HERMES \cite{askhermes}& \begin{tabular}[c]{@{}l@{}}Summary Answer\\ Generator\end{tabular} & Classical method & \begin{tabular}[c]{@{}l@{}}Preprocess questions to identify\\ keywords, followed by BM25\\ based retrieval.\end{tabular} \\ \hline
BioAnswerfinder \cite{bioanswerfinder20} & \begin{tabular}[c]{@{}l@{}}Snippet Retrieval \\and reranking\end{tabular} & Neural method & \begin{tabular}[c]{@{}l@{}}LSTM model identifies keywords.\\ Query expansion and search.\\ Reranking using BERT based\\ model.\end{tabular} \\ \hline
BERT \cite{bert} + BM25 & Document Retrieval & Neural Method & Hybrid method. \\ \hline
BM25 \cite{robertson2009probabilistic} + JPDRMM \cite{auebbioasq8} & \begin{tabular}[c]{@{}l@{}}Document Retrieval \\ and Reranking\end{tabular} & Neural Method & \begin{tabular}[c]{@{}l@{}}Retrieval using BM25,\\ Reranking using DRMM based\\ model.\end{tabular} \\ \hline
\end{tabular}%
}
\end{table}

\subsubsection{Classical (Lexical or Rule Based) Methods}

BM25 \cite{robertson2009probabilistic} is used very often for document retrieval for biomedical question answering, especially in the BioASQ challenge. It is used in its original form or with modifications or additional reranking steps. One reason for the success of BM25 based models in the BioASQ challenge lies in the construction of the dataset. While constructing the dataset, annotators search for keywords, identify the abstracts (which are ranked using standard TFIDF score) \cite{bioasq2015} and then frame the question. From the retrieved set of abstracts, the relevant ones are added to as gold standard context passages for the question. Consequently, the gold standard set of abstracts only have documents which have high TFIDF with some of the terms in the question. Other documents which may be relevant for answering the questions but are not lexically similar with the question, may be ignored. This is a cause of a bias in the gold standard document set which might make it difficult for other non sparse retriever models to perform well in the evaluation step, even if they actually retrieve relevant documents. However, several other methods have also been developed to retrieve or rerank passages for questions.

A method for document retrieval exploiting the UMLS semantic network was described in \cite{sarrouti2016generic}. The method can be divided into two steps. At first, the UMLS concepts are identified in the questions and converted into a disjunctive form, which is used as a query to search the Medline corpus. Then the retrieved documents are reranked by finding the semantic similarity \cite{mcinnes2009umls} between the question and the document title. In brief, this similarity depends on the path length (in the UMLS semantic network) between UMLS concepts present in the questions and the document titles. This methodology of document retrieval and reranking has been used in the BQA systems SemBioNLQA \cite{sembionlqa}.

SemRep Summaries \cite{semrepsum} can be used to generate semantic predications (subject-predicate-object relations) as the summary of a given abstract text. The extracted predications can be used to enchance document search to given questions. Query expansion has also been used for enhancing retrieval of abstracts. Essie \cite{essie} is an information retrieval tool with term and concept query expansion, and probabilistic ranking based on CQA \cite{cqa}. It uses a combination of knowledge-based and statistical techniques to find relevant abstracts. A study \cite{sneiderman2007} conducted to compare the mentioned methods with a baseline PubMed search engine showed that all the methods performed better than the baseline in finding relevant answers to questions asked by clinicians. Moreover, the study showed that the methods used in fusion (combination) leads to even better performance.

Structured knowledge can also help in the generation of "Ideal Answers" in the BioASQ Challenge \cite{bioasq2015}. Ideal Answers are typically descriptive answers which can summarize information present in several relevant documents. Kumar et. al. \cite{naresh-kumar-etal-2018-ontology} developed a QA system which involves finding relevant snippets from document exploiting a quasi ontology (knowledge graph) extracted from the documents. In brief, a rule based system is used to extract relations from PubMed abstracts, and a graph is constructed using them. Along with the relations, the text snippets from which relations are extracted are also stored. A Learning to Rank method is used to retrieve and rank the snippets according to the questions. Finally the snippets are summarized by a separate module to produce the final answer. The method produces better answer summaries which are more readable compared to other methods, demonstrated by a higher ROUGE score.

Constructing summary answers using different sentences containing non-redundant information has also been explored in other methods viz. OAQA \cite{oaqa17}, BioAMA \cite{bioama} and UNCC QA \cite{unccqa18}. OAQA \cite{oaqa17} describes a method for construction of summary answers using the gold standard abstracts snippets provided along with the question. In order to remove redundant sentences, a sentence similarity score based on word2vec \cite{word2vec} word embeddings, TFIDF cosine similarity as well as their original position in the list of abstracts provided. Using agglomerative clustering, clusters of similar sentences are identified. The final list of answer sentences is constructed using Maximum Marginal Relevance (MMR) \cite{mmr98}, which is a summarization technique that is able to tackle the issue of redundancy while maintaining question relevance. The final list of answer sentences are processed to produce the summary answers. For constructing summary answers, the BioAMA \cite{bioama} method breaks down the provided snippets into individual sentences and then selects the most relevant set of non redundant sentences by using a retrieval technique Indri \cite{indri}. Finally, it uses MMR \cite{mmr98} to select the relevant non redundant set of sentences and construct the answer. UNCC QA \cite{unccqa18} uses lexical chaining, a method for identifying semantically related words that represent the concept or the semantic meaning of a sentence, for quantifying sentence similarity and ranking.

AskHERMES \cite{askhermes} uses text from different sources viz Pubmed abstracts, full text PubMed Central articles, eMedicine documents, clinical guidelines and Wikipedia documents for answering questions with a summary type answer. The system processes documents to include information present in tables, remove short anchor texts and merge all section headers with the corresponding section body texts. For every question, AskHERMES identifies its type (device, diagnosis, history, management, pharmacology etc.) by using a SVM classifier. It also identifies the keywords using MetaMap \cite{metamap} and maps the keywords to UMLS concepts and semantic types \cite{umls2004}. Finally a BM25 \cite{robertson2009probabilistic} based model is used to retrieve relevant documents. From the retrieved abstracts, relevant passage are identified using a scoring function which incorporates word level and word similarity level similarity between sentences in the passage and keywords in the question. Finally the answers are hierarchically clustered based on the presence of keyword terms, processed to remove redundant information and presented to the user.

\subsubsection{Neural Methods}

Neural Networks have also been used for document retrieval, either for the initial retriever \cite{mahybrid}, or for reranking sentences \cite{bioanswerfinder20} or documents \cite{mahybrid, auebbioasq6, auebbioasq7, auebbioasq8}. Different combinations of retrieval and reranking in the context of biomedical question answering (BioASQ dataset) using pretrained language models like BERT \cite{bert} has been explored \cite{mahybrid}. It was shown that using a hybrid method incorporating a dual encoder BERT model along with BM25 model can improve document retrieval. This can be attributed to to BERT's ability to find semantically similar questions and documents and BM25 model ability to find lexically similar ones. It was also demonstrated that using a BERT cross attentional reranker can improve performance even more.

Neural network based reranker models viz. JPDRMM \cite{auebbioasq6, auebbioasq7, auebbioasq8} based on the DRMM architecture \cite{drmm} have consistently performed well in the BioASQ challenge. The entire retrieval system consists of a BM25 based retriever which retrieves the top 100 documents. These documents are then reranked using the JPDRMM model to find the 10 best documents.

Bio-Answerfinder \cite{bioanswerfinder19, bioanswerfinder20} is another method for finding answer sentences from PubMed abstracts. It uses several linguistic rules as well as a LSTM \cite{lstm} model to identify the key terms in the questions. The key terms are expanded in order to include different possible variations of the terms. Next, it iteratively relaxes the expanded queries and retrieves PubMed abstracts using the queries. In order to identify the answer sentences, the passages are broken down into individual sentences. Additionally the focus type of the question is also identified using linguistic rules and further used to filter the sentences. Another layer of filtering is done using a weighted-relaxed word mover's distance based similarity on word or phrase embeddings. Finally, a BERT \cite{bert} based reranker is used to rerank the filtered sentences and select the most relevant ones.

\subsection{Methods for Machine Reading}

\begin{table}[]
\centering
\caption{Methods for answering factoid or list based biomedical questions. There are several other methods based on similar strategies. They have been not discussed in this review.}
\resizebox{\textwidth}{!}{%
\begin{tabular}{llll}
\hline
\textbf{Method} & \textbf{Task} & \textbf{Method Classification} & \textbf{Remarks} \\ \hline
Lin \cite{lin2008} & Factoid Answer & Rule Based Method & \begin{tabular}[c]{@{}l@{}}uses Semantic Role\\ Labelling\end{tabular} \\ \hline
BioAnswerFinder \cite{bioanswerfinder20} & Factoid Answer & Rule Based Method &  \\ \hline
OAQA & Factoid & Machine Learning &  \\ \hline
LSTM based QA model \cite{neural_domain_adaptation} & Factoid, List & Deep Learning Model & \begin{tabular}[c]{@{}l@{}}Based on BioBERT \cite{biobert}; \\Pretrained on SQuAD 1.1 \cite{rajpurkarSQuAD1000002016} \end{tabular}\\ \hline
BERT + domain adaptation \cite{domain_portability} & Factoid, List & Deep Learning Model & \begin{tabular}[c]{@{}l@{}}Pretrained on Natural Questions \cite{kwiatkowski2019natural}; \end{tabular} \\ \hline
BioBERT(Yoon 7b) \cite{biobertbioasq7b} & Factoid, List & Deep Learning Model & \begin{tabular}[c]{@{}l@{}}Pretrained on SQuAD 1.1 \cite{rajpurkarSQuAD1000002016} \\ and SQuAD 2.0. \cite{rajpurkarKnowWhatYou2018} \end{tabular} \\ \hline
BioBERT(Jeong 8b) \cite{biobertbioasq8b} & Factoid, List & Deep Learning Model & \begin{tabular}[c]{@{}l@{}}Pretrained on SQuAD 1.1 \cite{rajpurkarSQuAD1000002016} \\ and SQuAD 2.0. \cite{rajpurkarKnowWhatYou2018}; \\ Pretraining on MedNLI task \cite{mednli} \end{tabular} \\ \hline
UNCC \cite{unccqa18} & Factoid, List & Deep Learning Model & Based on BioBERT \cite{biobert}; No Pretraining \\ \hline
\end{tabular}%
}
\end{table}

In an ideal end-to-end QA system aimed at predicting factoid answers, the document or passages retrieved by the retriever (and reranker) is used by another system (often called the Reader) to predict the final answers. Traditional readers used to depend on handcrafted features and processing steps to predict the answers. Also, different types of systems were designed for different answer types viz. yes/no, factoid, list and summary. Depending on the type of models used, the methods can be classified into two classes: 
\begin{enumerate}
    \item Non Neural Methods
    \item Neural Methods
\end{enumerate}

\subsubsection{Non Neural Methods}

These methods primarily uses hand crafted linguistic features and non neural machine learning models to predict the answers. For answering factoid or list type of questions, similar types of methods are usually applied. These methods very often use one or more the following steps using different types of techniques: 
\begin{enumerate}
    \item Identification the answer concept type ("protein", "gene", "drug" etc.) from the question
    
    \item Finding all candidate entities belonging to the answer concept type.
    
    \item Ranking of all candidates based on different linguistic rules and / or machine learning or learning to rank techniques.
    
\end{enumerate}

A method for answering factoid based on named entity recognition (NER) and semantic role labelling (SRL) was described in \cite{lin2008}. In brief NER and SRL is used to identify the entities and predicates in the question and hand crafted features are used to identify the answer entity type. The Queries are expanded with synonyms (using Wordnet \cite{wordnetlexical} and Longman's dictionary \cite{longmanlexicon}) and google search is use over a PubMed index to retrieve passages. NER and SRL is used to identify answer candidates, argument phrases and types, as well as whether answer candidates belong to the expected answer type. Finally, a linear ranking model ranks the answer candidates using the extracted features and the highest ranked answer candidate is predicted as the final answer. 

BioAnswerfinder \cite{bioanswerfinder20} predicts the exact or factoid answers by using an unsupervised rule based method from retrieved answer sentences. In brief, it identifies all entities or noun phrases in the retrieved answer sentences as candidates. These candidates answers are then filtered using word embedding (GloVe \cite{glove}) similarity among candidate answers and predicted answer types, removal of generic candidate answers ("gene", "protein" etc.) and removal of candidates present in the question. Finally the remaining answer candidates are ranked based on their TFIDF score and returned as the final prediction. Although this is an unsupervised technique, this works quite well on the BioASQ dataset and is one of the best performing methods. On the other hand, other methods using supervised machine learning techniques have also been developed, for e.g. OAQA \cite{oaqa17}. OAQA uses a logistic regression classifier to identify answer type using linguistic and grammatical features. Also, based on the type of the question, answer type and biomedical named entities present in the passages, candidate answers are identified. Finally the a logistic regression classifier ranks the candidate answers using several hand crafted features, and the top ranked answers are returned.

\subsubsection{Neural Methods}

Neural or deep learning models have been used for biomedical question answering on the BioASQ challenge for several years. Initially, the main focus was on LSTM \cite{lstm} based models \cite{neural_domain_adaptation}. Since the BioASQ datasets are quite small for training highly parameterized deep learning models, pretraining the QA models on the other QA datasets like SQuAD \cite{rajpurkarSQuAD1000002016} was done \cite{neural_domain_adaptation}. Gradually with the development of even more powerful deep learning architectures for NLP viz. transformer \cite{transformers} models, BQA reader systems increasingly started using them.

Currently, most BQA reader methods are based transformer based pretrained language models such as BERT \cite{bert}. Deep Learning or neural network based methods currently dominate the BioASQ challenge. Because of the success of transformer \cite{transformers} based language models in several general domain tasks \cite{bert, lewis-etal-2020-bart, 2020t5, roberta}, several language models pretrained on biomedical text have been explored \cite{biobert, biomedroberta, bioclinicalroberta, bioelmo, pubmedbert} for solving biomedical natural language processing problems. The BioBERT model \cite{biobert}, along with its variations, has consistently been one of the top performers in the BioASQ challenge in recent years \cite{biobertbioasq7b, biobertbioasq8b}.

BioBERT \cite{biobert} is a pretrained language model designed for biomedical NLP tasks such as entity recognition, question answering, information extraction etc. It is similar to the original BERT \cite{bert} model, with the different being that BioBERT was pretrained specifically using biomedical text. For the different tasks that BioBERT can perform, the pretrained model architecture needs to be slightly changed and the model needs to be fine tuned for the specific task. For factoid biomedical question answering, BioBERT can be modified to predict a span of the text as the answer \cite{biobertbioasq7b, biobertbioasq8b}. In brief, the question and a context paragraph is concatenated and input into a BioBERT model. The model predicts two positions in the text, one denoting the start and end positions of the predicted answer. The model can be trained with several training examples of question-context pairs and known answer span positions. Since transformer based models are essentially data hungry and need a large amount of training data, the relatively small dataset provided with BioASQ challenge \cite{bioasq2015} is not enough for training the model. Models based on BioBERT \cite{biobert} or the original BERT \cite{bert} have been pretrained by different research groups using a variety of pretraining techniques, including training on general domain QA datasets like SQuAD \cite{rajpurkarSQuAD1000002016, biobertbioasq7b} and Natural Questions \cite{kwiatkowski2019natural, domain_portability} before being fine tuned with the much smaller BioASQ data \cite{biobertbioasq7b}.

The BioBERT QA model \cite{biobertbioasq7b} can be further improved by using innovative ways of pretraining the model, which enables the model to learn to reason and make inferences \cite{biobertbioasq8b}. After pretraining the BioBERT QA model with the SQuAD data, the models are further trained for medical natural language inference task using the MedNLI dataset \cite{mednli}. Finally training on the BioASQ dataset is carried out. It was demonstrated that addition of the MedNLI task to the pretraining regime improved QA performance, a possible reason being the better understanding ability of the model due to the additional training step. Additionally, it was also seen that fusing external features (like part of speech tags, named entities etc.) along with the usual the BioBERT model \cite{externalbqa} can improve the accuracy of the model even more. Other ways for pretraining which has been shown to improve the QA performance of the BioBERT model include predicting spans in the text where original named entities were replaced with new ones \cite{unccqa18}.

Although the BioBERT QA models \cite{biobertbioasq7b, biobertbioasq8b} have mostly outperformed other models on the BioASQ tasks, they still suffer from certain drawbacks. There are several questions in the BioASQ data in which the answers do not belong to any span in the context or passage text. Therefore, BERT based extractive QA models like BioBERT cannot predict those answers correctly for these questions. In the general domain, generative transformer models built on BART \cite{lewis-etal-2020-bart} and T5 \cite{2020t5} have outperformed the extractive models on QA tasks \cite{lewisRetrievalAugmentedGenerationKnowledgeIntensive, izacardLeveragingPassageRetrieval2020}. Therefore, one potential solution to the problem of answers not belonging to a text span can be solved by pretrained genarative transformer based language models, at least in theory. However, whereas pretrained BERT \cite{bert}, ElMo \cite{elmo} and RoBERTa \cite{roberta} models for the biomedical domain are available publicly \cite{biobert, biomedroberta, bioclinicalroberta, bioelmo, pubmedbert}, similar pretrained biomedical generative language models like BART \cite{lewis-etal-2020-bart} and T5 \cite{2020t5} for the biomedical domain are not available publicly. Since training these genarative models is a computate intensive process, training them is a substantial bottleneck. Another drawback of the BioBERT QA models \cite{biobertbioasq7b, biobertbioasq8b} is its assumption that only relevant abstracts will be provided with the questions. This assumption can lead to problems, especially when used as a reader in a retriver-reader end-to-end (open domain) question answering system. Even the state-of-the-art retrievers will often provide several noisy abstracts, and the readers should be able to predict the answers correctly even when the set of abstracts provided to it contains a irrelevant ones. However, it was found that adding a few irrelevant abstracts to a set of relevant abstracts or removing a few relevant ones substantially reduces the accuracy of the BioBERT QA model. Therefore, more robust BioBERT QA models which are able to predict correct answers even when noisy abstracts are provided are needed, something that can only be achieved using end to end training of retrievers and readers.

\section{Challenges and Future Directions} \label{DISCUSS}

\subsection{Challenges of Biomedical QA systems}
Despte the progress made by biomedical question answering systems, there are significant challenges that still remain \cite{biomedicalqareview}:

\subsubsection{ Lack of Large Datasets} Since most methods currently used for BQA are based on deep learning, they are extremely data hungry. However, development of benchmark BQA datasets require manual annotation by medical and biological experts, which is an expensive and time consuming exercise. Therefore, whereas large QA dataset like SQuAD \cite{rajpurkarSQuAD1000002016, rajpurkarKnowWhatYou2018} and Natural Questions \cite{kwiatkowski2019natural} are available for the general domain, most of manually annotated BQA datasets are small compared to their general domain counterparts. BioASQ, one of the largest BQA datasets has close to four thousand questions, which is miniscule compared to SQuAD (more than 100000) and Natural Questions (around 80000). This gap, coupled with the data hungry nature of current state of the art deep learning models have led to the development of datasets which are constructed using automated methods. However, their reliability and utility are limited.
    
\subsubsection{ Limited use of Knowledge Bases} Most methods for BQA developed recently aim at answering questions using text. In the process, they fail to utilize the rich biomedical resources viz. especially biomedical knowledge bases. It should be mentioned that there have been some attempts to encourage the development of KGQA methods for the biomedical domain, most important among them being the QALD4 Task 2 Challenge \cite{qald4}. However, the challenge focuses on a small subdomain of biomedicine, viz. the relations among drugs treatments with diseases and side effects, and therefore narrow in scis very ope. Another issue is that knowledge in the domain is scattered across different such KBs and hence collating them into a knowledge graph with a unified schema may be difficult.
    
\subsubsection{ Lack of Explainability and Interpretability} 
BQA is a complicated process and can potentially determine clinical decision making. Therefore, it is important for BQA methods to provide the explanations for predicting a particular answer apart from the answer itself, so a human can judge the answer and reason if it is reliable enough. However, most methods focus on predicting the answer and ignore explaining them.
    
\subsubsection{ Fairness and Bias} Biomedical scientific literature and electronic health records are two resources commonly used by biomedical QA systems for learning the QA models as well as answering the questions. However, in some cases, the information present in the texts can be outdated, which is not taken into account by most methods for BQA.
    
\subsubsection{ Focus on Machine Comprehension} With the development of BioASQ challenge, in recent years, it can be seen that the participants have mostly chosen to focus on one task: machine comprehension. However, effectiveness of end-to-end question answering systems also depend on several other components like information retrieval and utilization of structured knowledge. Therefore, end-to-end BQA systems still lack the accuracy and efficiency to be of practical use. 
    
\subsubsection{ Weakness of Evaluation Techniques} Due to the variety of jargon viz, the rich synonym relationships, used in biomedicine, specialized methods for evaluation of BQA systems are necessary. However, the currently used evaluation techniques, especially for BioASQ, are naive and fail to take this variety into consideration.
    

\subsection{Future Directions}

Question answering in the domain of biology has a lot of scope for growth and has several applications in real life clinical and scientific settings. In order to improve biomedical QA systems and make it a practical utility for clinicians, scientists as well as consumers, the following research paths will need to be explored or exploited.

\subsubsection{Integration of Structured Knowledge and Text} 
As discussed several times in this review, QA using knowledge bases and text have their own set of advantages and disadvantages. Highly precise knowledge bases and reliable scientific text are available in the domain of biomedicine. Currently, in several BQA systems, both text and knowledge bases (in particular UMLS thesaurus) are used, however, in most cases such integration is superficial (primarily used for named entity recognition) \cite{sembionlqa, abacha2015means}. In the future, we expect that integration of different structured and unstructured sources of knowledge and information would increase and consequently lead to improved biomedical QA systems.
    
\subsubsection{More focus on End-to-End QA Pipelines}
In recent years because of the popularity of the BioASQ challenge, there has been a rise in information retrieval (IR) and machine reading comprehension systems (MRC) which work in isolation. In the general domain however, a lot of improvement in ODQA accuracy can be seen by end-to-end systems in which the IR and MRC components are trained jointly \cite{izacardDistillingKnowledgeReader2020, lewisRetrievalAugmentedGenerationKnowledgeIntensive, karpukhin-etal-2020-dense}. We expect jointly optimized BQA systems to outperform systems with components trained in isolation.
    
\subsubsection{Transfer Learning, MultiTask Learning and Meta Learning}
It has already been demonstrated that pretrained models work well in answering biomedical questions \cite{biobertbioasq7b, biobertbioasq8b}. Moreover, it was also shown that models trained on other related tasks such as natural language inference can answer questions better \cite{biobertbioasq8b}. In the general domain, the current state of the art ODQA model, FiD \cite{izacardLeveragingPassageRetrieval2020} is based on T5 model \cite{2020t5}. T5 itself is trained on multiple tasks and commonly used for transfer learning. Pretraining transformer models on different biomedical NLP tasks such as information extraction, named entity recognition, natural language inference etc. can improve BQA performance. Therefore, exploring multitask learning and transfer learning seems to be a good track for research to improve BQA systems.
    
\subsubsection{Explanation and Interpretation of Answers}
Ever since the boom in deep learning technologies, the main focus was to develop black box systems with high predictive power on a given task. It was seen that although deep learning based black box models outperform other methods if provided with enough data and compute power, such systems are often easily fooled by adversarial attacks. This vulnerability to adversarial attacks coupled with the lack of explanations of predictions has led to a growth in interest in explainable AI systems. Explainability and interpretabilty is a significant part of machine learning in the field of biomedicine \cite{amann2020explainability, holzingerexplain, interpretationbmml}, since humans will not be able to trust a system which cannot provide a reason for providing a particular answer. Therefore, we expect more research to focus on not just predicting answers but providing an explanation to its prediction as well.

\subsubsection{More Datasets}
Lack of large datasets is a significant cause of poor performance of biomedical QA system to its general domain counterparts. However, several new datasets constructed from medical or biological examination questions have been described \cite{li2020nlpec, vilares2019head}. This is a promising direction, since it can produce a significant number of question-answer pairs while limiting the expenses. However, most of such datasets comprise of multiple choice question and may not be used directly for open domain question answering. Another drawback of this methodology is that datasets might be in different languages e.g. NLPEC \cite{li2020nlpec}, and translating them into a common language can be difficult.
    
\subsubsection{Conversational Systems}
Conversational systems have shown significant improvements \cite{coqa, coqareview} and have been adopted by several industries, primarily for customer support. Since there is a significant shortage of doctors, especially in developing countries, whereas an significant fraction of the population have access to smart phones and internet \cite{smartphones}, having biomedical conversational systems can help patients get medical advice.
    
\subsubsection{Inference Based Systems}
The field of biomedical informatics has shown significant development in recent years. Machine learning, especially deep learning and graph based machine learning technologies have helped in several biomedical predictive tasks, such as predicting drug target interactions \cite{deeppurpose}, drug adverse effects \cite{sideeffectsdrug}, prediting diseases from symptoms \cite{NEURIPS2020_5bca8566}, etc \cite{li2021representation}. Additionally, in the general domain, several techniques for reasoning and querying over knowledge graphs have been developed \cite{NEURIPS2018_ef50c335, lin-etal-2018-multi-hop,
guo-etal-2016-jointly, xiong-etal-2017-deeppath, guu-etal-2015-traversing, yang2014embedding, das2017chains}. Therefore, one interesting field of research would be to provide biomedical QA systems the ability to reason and predict completely novel answers to natural language biomedical questions.

\subsubsection{Regulation}: Since biomedical question answering systems may have an effect in medical decision making and therefore may influence human survival, much like autonomous vehicles, they are expected to be regulated. Therefore, the main focus should not just be on developing highly accurate BQA systems, but also on making them incorruptible, reliable and transparent.

\section{Conclusion}
In this review, we focused on the recent advances in biomedical question answering. We provided a review of general domain question answering using knowledge bases, texts or both, before moving on to biomedical question answering systems. We explored current state of the art biomedical QA systems and discussed their limitations. We finally provided an overall analysis of limitations of BQA systems along with ways to overcome them. We finally explore the potential areas of focus for further research. 
\bibliographystyle{plain}
\bibliography{citations}
\end{document}